\documentclass[nohyperref]{article}
\usepackage[accepted]{icml2022}
\usepackage[utf8]{inputenc}
\usepackage[T1]{fontenc}    
\usepackage[hidelinks]{hyperref}       
\usepackage{booktabs}       
\usepackage{amsfonts}       
\usepackage{amsmath}        %
\usepackage{nicefrac}       
\usepackage{algorithm,algorithmic}
\usepackage{colortbl}
\usepackage{cleveref} 
\crefname{equation}{}{}

\definecolor{mygray}{gray}{0.8}
\usepackage[colorinlistoftodos]{todonotes}
\usepackage{color, mathtools, enumerate}
\usepackage{xcolor}
\usepackage{colortbl}
\usepackage{multirow}
\usepackage{threeparttable}
\usepackage{appendix}
\usepackage{rotating}
\usepackage[export]{adjustbox}

\def\R{\mathbb{R}}

\def\D{\mathcal{D}}

\newcommand{\norm}[1]{\left\|#1\right\|}

\def\argmax{\mathop{\rm arg\,max}\limits}
\def\argmin{\mathop{\rm arg\,min}\limits}
\def\minop{\mathop{\rm min}\limits}
\def\maxop{\mathop{\rm max}\limits}

\def\min{\mathop{\rm min}\nolimits}
\def\max{\mathop{\rm max}\nolimits}

\def\xorig{\boldsymbol{x}_\textrm{orig}}
\def\xadv{\boldsymbol{x}_\textrm{adv}}

\newcommand{\iter}[2]{#1^{(#2)}}
\newcommand{\bs}{\boldsymbol}

\usepackage{color}
\usepackage{comment}

\newcommand{\measurement}{DI~}
\newcommand{\measurementNoSpace}{DI}
\newcommand{\measurementFirst}{Diversity Index~}
\newcommand{\measurementFirstNoSpace}{Diversity Index}

\hypersetup{draft}
\begin{document}
\twocolumn[
\icmltitle{Diversified Adversarial Attacks based on Conjugate Gradient Method}



\begin{icmlauthorlist}
\icmlauthor{Keiichiro Yamamura}{yyy}
\icmlauthor{Haruki Sato}{yyy}
\icmlauthor{Nariaki Tateiwa}{yyy,ntt}
\icmlauthor{Nozomi Hata}{comp}
\icmlauthor{Toru Mitsutake}{yyy}
\icmlauthor{Issa Oe}{yyy}
\icmlauthor{Hiroki Ishikura}{yyy}
\icmlauthor{Katsuki Fujisawa}{comp}
\end{icmlauthorlist}

\icmlaffiliation{yyy}{Graduate School of Mathematics, Kyushu University, Fukuoka, Japan.}
\icmlaffiliation{comp}{Institute of Mathematics for Industry, Kyushu University, Fukuoka, Japan.}
\icmlaffiliation{ntt}{Present affiliation is NTT Software Innovation Center, NTT Corporation.}

\icmlcorrespondingauthor{Keiichiro Yamamura}{keiichiro.yamamura@kyudai.jp}

\icmlkeywords{Machine Learning, ICML}

\vskip 0.3in
]
\printAffiliationsAndNotice{}

\begin{abstract}
Deep learning models are vulnerable to adversarial examples, and adversarial attacks used to generate such examples have attracted considerable research interest.
Although existing methods based on the steepest descent have achieved high attack success rates, ill-conditioned problems occasionally reduce their performance.
To address this limitation, we utilize the conjugate gradient (CG) method, which is effective for this type of problem, and propose a novel attack algorithm inspired by the CG method, named the Auto Conjugate Gradient (ACG) attack.
The results of large-scale evaluation experiments conducted on the latest robust models show that, for most models, ACG was able to find more adversarial examples with fewer iterations than the existing SOTA algorithm Auto-PGD (APGD).
We investigated the difference in search performance between ACG and APGD in terms of diversification and intensification, and define a measure called Diversity Index (DI) to quantify the degree of diversity.
From the analysis of the diversity using this index, we show that the more diverse search of the proposed method remarkably improves its attack success rate.
\end{abstract}
\begin{figure}[tb]
    \centering
    \includegraphics[width=\linewidth]{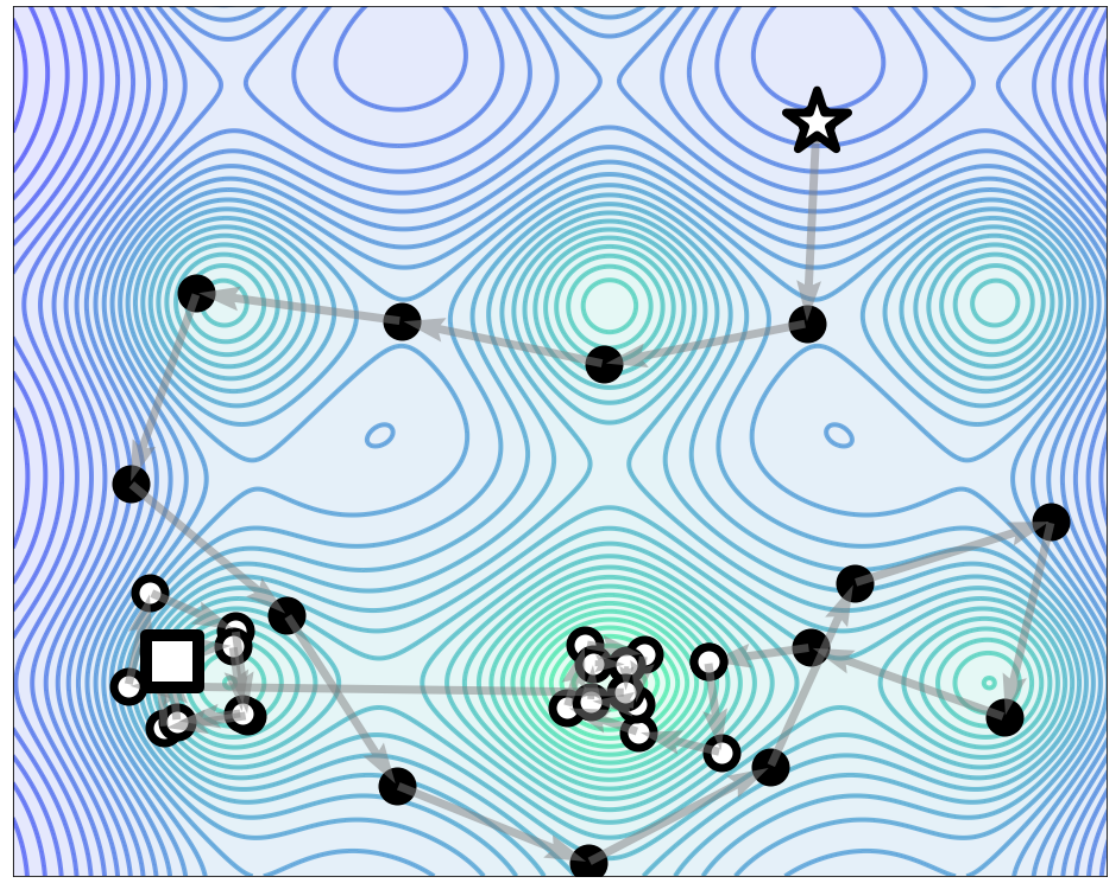}
    \caption{
    An example of our intended search for a multimodal function.
    The search finds the global optima owing to an appropriate balance between diversification and intensification.
    The initial search point is indicated by the white star, and the search ends at the white square. 
    The circles represent the search points, and the
    black circles indicate that the search has been diversified.
    }
    \label{fig:DiversificationAndIntensification}
    \vskip -0.2in
\end{figure}
\section{Introduction}
Deep learning models are effective for various machine learning tasks, and are increasingly being applied to safety-critical tasks such as automated driving.
However, deep learning models may misclassify \emph{adversarial examples}~\cite{szegedy2013intriguing,goodfellow2014explaining} formed by applying perturbations to their inputs which are too small for the human eye to perceive.
Hence, improving the robustness of deep learning models against adversarial examples is crucial for safety-critical tasks.
\emph{Adversarial training} \cite{goodfellow2014explaining} is an effective method used to create robust models.
In adversarial training, adversarial examples are generated and added to the training data. This requires many adversarial examples to be generated quickly for efficient training.

An \emph{adversarial attack} is a method used to generate adversarial examples.
A \emph{white-box adversarial attack} assumes that the algorithm can obtain the outputs and gradients of the deep learning models.

The fast gradient sign method (FGSM) \cite{goodfellow2014explaining}, iterative-FGSM (I-FGSM) \cite{I-FGSM}, and projected gradient descent (PGD) attack \cite{madry2018towards} use the sign of the steepest gradients at the update.
In addition, some attack strategies have successfully improved on the performance of these methods by introducing past search information.
Momentum I-FGSM \cite{dong2018boosting} and Nesterov I-FGSM \cite{Lin2020Nesterov} are based on the momentum method, which simultaneously considers the current and past gradient to determine the next update.
Similarly, Auto-PGD (APGD) \cite{croce2020reliable}, based on projected gradient descent (PGD) also considers the inertia direction of the search point in the next update.
However, searches performed by algorithms based on the steepest gradient descent may be insufficient because the objective function of an adversarial attack is nonconvex, nonlinear, and multimodal.
We classify the existing methods according to their settings in \Cref{appendix:related_works}.
The relevant literature on adversarial examples has been summarized at the URL provided below\footnote{\url{https://nicholas.carlini.com/writing/2019/all-adversarial-example-papers.html}}.

In this study, a new white-box adversarial attack called an Auto Conjugate Gradient (ACG) attack is proposed based on the conjugate gradient (CG) method.
The CG method is a well-known algorithm for systems of linear equations and is also used in nonlinear optimization. It updates the search point in more diverse directions compared to the steepest direction, and can be predicted to search extensively (see \Cref{fig:DiversificationAndIntensification} and \Cref{appendix:exampleDiversificationAndIntensification}). 

To the best of our knowledge, our proposed method is the first adversarial attack with a high performance based on the CG.
We compared our ACG with APGD, a SOTA white-box adversarial attack, on 64 robust models listed in RobustBench~\cite{croce2020robustbench}.
The results show that the attack success rate (ASR) of ACG was much higher than that of APGD, with the exception of only a single model (see \Cref{table:cifar10_large,table:cifar100_large}).
Surprisingly, ACG with 1 restart (100 iterations) performed better than APGD with 5 restarts (5 $\times$ 100 iterations) against approximately three-fourths of all robust models, although the execution time per iteration of ACG and APGD was almost equal.
We thoroughly analyzed the factors involved in the improved attack performance of ACG.
Compared to APGD, it may be observed from the results that the movement of the search points in ACG was large, and the attacked class was varied more often during the search.

Moreover, we analyze ACG and APGD for \emph{diversification and intensification}.
Diversification and intensification have received considerable attention in the field of metaheuristics \cite{CrepinsekMatej2013}, where the objective function is generally nonconvex and multimodal, similar to deep neural networks.
To control the algorithms properly, some studies on metaheuristics~\cite{Cheng2014,Morales-Castaneda2020} 
have attempted to quantify the balance between diversification and intensification.
However, to the best of our knowledge, no such methods have been proposed for gradient-based iterative searches such as adversarial attacks. 
Therefore, we propose \emph{\measurementFirstNoSpace} (\measurementNoSpace) to quantify the degree of the diversity of the search points and analyze adversarial attacks (see \Cref{subsection:measurement}).
Compared to APGD, we demonstrate that ACG can search more extensively by an analysis of the \measurement.

The contributions of this study are summarized as follows.
\begin{itemize}
    \item We propose a new adversarial attack called ACG. In a large-scale experiment on 64 robust models, the ASR of ACG overwhelmingly outperformed that of APGD, a SOTA adversarial attack, except for a single model (see \Cref{sec:experiments}).
    The ASR of ACG with 1 restart (100 iterations) is generally better than that of APGD with 5 restarts (5 $\times$ 100 iterations).
    \item We propose a metric \measurement to quantify the degree of diversity and intensity of the search points of gradient-based iterative search algorithms.
    The \measurement measure was evaluated, and the results indicated that the search performed by ACG was more diversified than that of APGD (see \Cref{section:studyOfACG}).
\end{itemize}
\vspace{-1.0\baselineskip}
Our code is available at the URL given below\footnote{\url{https://github.com/yamamura-k/ACG}}. 
\section{Preliminaries}
\label{preliminaries}

\subsection{Problem Settings}
Let the locally differentiable function $g:\D \subseteq \R^m \to \R^K$ be a $K$-classifier that classifies by $\argmax_{k=1,\ldots, K}(g_k(\cdot))$, and let $\xorig \in \D$ be a point classified as class $c$ by $g$.
Given the distance function $d(\cdot, \cdot)$ and $\varepsilon >0 $, the feasible region $\mathcal{S}$ in an adversarial attack is defined as $\mathcal{S}=\{\bs{x} \in \D  \,|\, d(\xorig,\bs{x})\leq \varepsilon\}$.
We then define an adversarial example as $\xadv \in \D$, which satisfies
\begin{align}
    \argmax_{k=1,\ldots, K}g_k(\xadv) \neq c, \ d(\xorig, \xadv)\leq \varepsilon.
\end{align}
Let $L$ be the objective function to search for $\xadv$. 
The adversarial attack can be formulated as follows. 
\begin{equation} 
\maxop_{\bs{x}\in\D} L(g(\bs{x}), c) \quad \textrm{s.t.} \quad d(\xorig, \bs{x})\leq \varepsilon .
\label{eq:formulation_adex}
\end{equation}
The above formulation renders $\bs{x}$ less discriminative to the class $c$ by $g$. 
In classifiers that apply image classification, the Euclidean distance $d(\bs{v},\bs{w}):=\norm{\bs{v} - \bs{w}}_2$, the uniform distance $d(\bs{v},\bs{w}):=\norm{\bs{v}-\bs{w}}_\infty$, and $\mathcal{D}=[0,1]^m$ are often used.
We refer to adversarial attacks that use the uniform distance as $l_\infty$ attack.

\subsection{PGD Attack}
The PGD method is effective for solving the problem \eqref{eq:formulation_adex}.
Given $f:\R^m \to \R$ and the formulation $\maxop_{\bs{x} \in \mathcal{S}}\; f(\bs{x})$, let the iterations in PGD be $k=1,\ldots,N_\textrm{ iter}$, where $\iter{\bs{x}}{k+1}=P_{\mathcal{S}}(\iter{\bs{x}}{k} + \iter{\eta}{k}\nabla f (\iter{\bs{x}}{k}))$, in which $\iter{\eta}{k}$ is the step size
and $P_{\mathcal{S}}$ is the projection onto the feasible region $\mathcal{S}$. 
APGD adds a momentum update method to PGD. 
Let $\iter{\bs{\delta}}{k}$ be the update direction for each iteration (e.g., $\nabla f(\iter{\bs{x}}{k})$ for a uniform distance case).
The update rules of APGD in a single iteration, including the momentum term, are defined as follows.
\begin{align}
    \iter{\bs{z}'}{k+1} =& \iter{\bs{x}}{k} + \iter{\eta}{k} \sigma(\iter{\bs{\delta}}{k})\\
    \iter{\bs{z}}{k+1} =& P_\mathcal{S}(\iter{\bs{z}'}{k})\\
    \iter{\bs{x}}{k+1} =& P_\mathcal{S}(\iter{\bs{x}}{k} + \alpha (\iter{\bs{z}}{k+1} - \iter{\bs{x}}{k}) +\\ &(1-\alpha)(\iter{\bs{x}}{k} - \iter{\bs{x}}{k-1})) \nonumber,
\end{align}
where $\sigma(\cdot)$ is a type of normalization and $\alpha$ is a coefficient representing the strength of the momentum term, and $\alpha=0.75$ is used in APGD.
\subsection{General Conjugate Gradient Method}
The conjugate gradient method (CG) was developed to solve linear equations and subsequently extended to the minimization of strictly convex quadratic and general nonlinear functions. 
Most existing works on CG methods have considered unconstrained optimization problems, but CG can be applied to constrained problems by using projection. 
Given an initial point $\iter{\bs{x}}{0}$, the initial conjugate gradient $\iter{\bs{s}}{0}$ is set to $\mathbf{0}$, and the $k$-th search point $\iter{\bs{x}}{k}$ and conjugate gradient $\iter{\bs{s}}{k}$ are updated as 
$    \iter{\bs{s}}{k} =-\nabla f(\iter{\bs{x}}{k})+\iter{\beta}{k} \iter{\bs{s}}{k-1},
    \iter{\eta}{k} =  \argmin \{ f(\iter{\bs{x}}{k} + \eta\iter{\bs{s}}{k}) \mid \eta \ge 0\}, \iter{\bs{x}}{k+1} = \iter{\bs{x}}{k} + \iter{\eta}{k}\iter{\bs{s}}{k}$
where $k \geq 1$ and $\iter{\beta}{k}$ is a parameter calculated from past search information.
The step size $\iter{\eta}{k}$ is usually determined by a linear search to satisfy some conditions such as the Wolfe conditions, because solving $\argmin \{ f(\iter{\bs{x}}{k} + \eta\iter{\bs{s}}{k}) \mid \eta \ge 0\}$ exactly is difficult.

Consider the problem of minimizing the strictly convex quadratic function
$\displaystyle f(x)=\bs{x}^T A\bs{x}+\bs{b}^T \bs{x}$, where $A$ is a positive definite matrix and $\bs{x} \in \R^n$. In this case, the coefficient $\iter{\beta}{k}$ is $\frac{\langle A\iter{\bs{s}}{k-1},\iter{-\nabla f(\bs{x}}{k})\rangle}{\langle A\iter{\bs{s}}{k-1},\iter{\bs{s}}{k-1}\rangle}$. 
When the objective function is a strictly convex quadratic function, CG is known to be able to find the global solution in less than $n$ iterations under an exact linear search.

For nonlinear functions, some formulas have been proposed to calculate $\iter{\beta}{k}$ have been proposed (for further details, see \cite{Hager2006}).
In this study, we use the following formula for $\iter{\beta}{k}$ proposed by \cite{Hestenes1952}, which exhibited the highest ASR in our preliminary experiments (\Cref{appendix:OtherConjugateFormulation}).
\begin{align}\label{eq:_beta_HS}
    & \iter{\beta}{k}_{HS}=\frac{\langle\nabla f(\iter{\bs{x}}{k}),\iter{\bs{y}}{k-1}\rangle}{\langle\iter{\bs{s}}{k-1}, \iter{\bs{y}}{k-1}\rangle},
\end{align}
where $\displaystyle \iter{\bs{y}}{k-1}=\nabla f(\iter{\bs{x}}{k})-\nabla f(\iter{\bs{x}}{k-1})$.
Theoretical convergence studies on CG methods often assume that $\iter{\beta}{k} \ge 0$ \cite{Hager2006} and some implementations use $\max \{\iter{\beta}{k}, 0\}$ instead of $\iter{\beta}{k}$.
\newif\ifreducealgo
\reducealgotrue

\begin{figure}[H]
    \centering
    \begin{algorithm}[H]
     \caption{ACG}
     \label{algo:method}
     \begin{algorithmic}[1]
     \STATE {\bfseries Input:}$f, \mathcal{S}, \iter{\bs{x}}{0}, \iter{\eta}{0}, N_{{\rm iter}}, W=\left\{w_0,\ldots,w_n\right\}$
     \STATE {\bfseries Output:} $\bs{x}_{{\rm adv}}$
\ifreducealgo
     \STATE $\bs{x}_{{\rm adv}} \leftarrow \iter{\bs{x}}{0}$; $\iter{\beta}{0} \gets 0$; $\iter{\bs{s}}{0} \gets \nabla f(\iter{\bs{x}}{0})$
     \STATE $\bs{x}_{\rm pre} \gets \iter{\bs{x}}{0}$; $\bs{s}_{\rm pre} \gets \iter{\bs{s}}{0}$
\else
     \STATE $\bs{x}_{{\rm adv}} \leftarrow \iter{\bs{x}}{0}$; $\bs{x}_{\rm pre} \gets \iter{\bs{x}}{0}$
\fi
     \FOR{$k=0$ {\bfseries to} $N_{{\rm iter}}-1$}
\ifreducealgo
\else
     \IF {$k = 0$} 
     \STATE $\iter{\beta}{0} \leftarrow 0$; $\iter{\bs{s}}{0}\leftarrow\nabla f(\iter{\bs{x}}{0})$; $\bs{s}_{\rm pre} \gets \iter{\bs{s}}{0}$
     \ELSE
     \STATE Compute $\iter{\beta}{k}$ \eqref{eq:beta_HS} and $\iter{\bs{s}}{k}$ \eqref{eq:update_sk}
     \ENDIF
\fi
     \STATE Compute $\iter{\bs{x}}{k+1}$     \eqref{eq:update_xk}
     \IF{$f(\iter{\bs{x}}{k+1})>f(\bs{x}_{{\rm adv}})$}
     \STATE $\bs{x}_{{\rm adv}} \gets  \iter{\bs{x}}{k+1}$; $\bs{x}_{\rm pre} \gets \iter{\bs{x}}{k}$; $\bs{s}_{\rm pre} \gets \iter{\bs{s}}{k}$
     \ENDIF
     \STATE $\iter{\eta}{k+1} \gets \iter{\eta}{k}$
     \IF {$k\in W$}
     \IF {Condition (\ref{eq:condition_1}) {\bfseries or}  (\ref{eq:condition_2}) is satisfied}
     \STATE $\iter{\eta}{k+1} \gets \iter{\eta}{k} / 2$;
     \STATE $\iter{\bs{x}}{k+1} \gets  \bs{x}_{{\rm adv}}$; $\iter{\bs{x}}{k} \gets \bs{x}_{\rm pre}$;  $\iter{\bs{s}}{k} \gets \bs{s}_{\rm pre}$
     \ENDIF
     \ENDIF
\ifreducealgo
     \STATE Compute $\iter{\beta}{k+1}$ \eqref{eq:beta_HS} and $\iter{\bs{s}}{k+1}$ \eqref{eq:update_sk}
\else
\fi
     \ENDFOR
     \end{algorithmic}
    \end{algorithm}
    \vskip -0.3in
\end{figure}
\section{Auto Conjugate Gradient (ACG) Attack}
\label{section:acg}
We propose the Auto Conjugate Gradient (ACG) attack as a novel adversarial attack inspired by the CG approach.
The proposed scheme is summarized in \Cref{algo:method}.

The major differences among ACG, general CG, and APGD are summarized in \Cref{table:differenceAcgGcgAPGD}. We apply the step size strategy employed in APGD instead of a linear search because the forward propagation is relatively time-consuming. In addition, we do not restrict $ \beta \geq 0 $, whereas the general CG usually makes $\beta$ non-negative (see \Cref{appendix:positivebeta_conjugate}).

\subsection{ACG Step}
To solve the maximization problem as an adversarial attack, we use $-\nabla f(\cdot)$ instead of $\nabla f(\cdot)$ in \Cref{eq:_beta_HS}, that is, 
\begin{align}
    \label{eq:update_y} \iter{\bs{y}}{k-1}&=\nabla f(\iter{\bs{x}}{k-1})-\nabla f(\iter{\bs{x}}{k}),\\
    \label{eq:beta_HS}\iter{\beta}{k}_{HS}&=
    \frac{\langle -\nabla f(\iter{\bs{x}}{k}), \iter{\bs{y}}{k-1} \rangle}
    {\langle \iter{\bs{s}}{k-1}, \iter{\bs{y}}{k-1} \rangle},\\
    \label{eq:update_sk}\iter{\bs{s}}{k} &=\nabla f(\iter{\bs{x}}{k})+\iter{\beta}{k}_{HS} \iter{\bs{s}}{k-1},\\
    \label{eq:update_xk}\iter{\bs{x}}{k+1} &= P_{\mathcal{S}}\left(\iter{\bs{x}}{k} + \iter{\eta}{k}\cdot\sigma(\iter{\bs{s}}{k})\right),  
\end{align}
where $\sigma(\cdot)$ is a type of normalization.
ACG uses a sign function as $\sigma$ because many previous studies including APGD use it for an $l_\infty$ attack.
There is a possibility of division by zero when calculating $\iter{\beta_{HS}}{k}$, and empirically, it occurs when $\iter{\bs{y}}{k-1}=\boldsymbol{0}$.
Therefore, when division by zero is called for, the issue is addressed by setting $\iter{\beta}{k}=0$.

\begin{table}[tb]
\vskip -0.1in
\caption{
The differences among ACG, General CG, and APGD.
$\iter{\bs{\delta}}{k}, \sigma, \iter{\eta}{k}$ refer to the update direction, whether to normalize $\iter{\bs{\delta}}{k}$, and the step size strategy, respectively (described in \Cref{preliminaries}). $\iter{\bs{s}}{k}$ is the CG direction. The "momentum" column refers to whether a momentum term is used in the update.
}
\label{table:differenceAcgGcgAPGD}
\vskip 0.1in
\centering
\begin{small}
\begin{tabular}{c|cccc}
    \toprule
           & $\iter{\bs{\delta}}{k}$ & momentum & $\sigma$ & $\iter{\eta}{k}$  \\ \midrule
ACG        & $\iter{\bs{s}}{k}$ & - &  \checkmark & \Cref{sec:Step Size Selection} \\
General CG &$\iter{\bs{s}}{k}$ & - &  -& linear search      \\
APGD       & $\nabla f(\iter{\bs{x}}{k})$ & \checkmark & \checkmark  & \Cref{sec:Step Size Selection}\\
\bottomrule
\end{tabular}
\end{small}
\vskip -0.1in
\end{table}

\subsection{Step Size Selection}
\label{sec:Step Size Selection}
We use the same method proposed in APGD to select the step size. 
The initial step size $\iter{\eta}{0}$ is set to $2\varepsilon$, and when the number of iterations reaches the precomputed checkpoint $w_j$, the step size $\eta$ is halved if either of the following two conditions are satisfied. 
\vspace{-0.5\baselineskip}
\begin{enumerate}[(I)]
    \setlength{\parskip}{0cm}
    \setlength{\itemsep}{0cm}
    \item $\label{eq:condition_1}\displaystyle N_{\rm inc} < \rho\cdot(w_j-w_{j-1})$,
    \item $\label{eq:condition_2}\displaystyle\iter{\eta}{w_{j-1}} = \iter{\eta}{w_j} \text{ and } \iter{f_{\max}}{w_{j-1}} = \iter{f_{\max}}{w_{j}}$,
\end{enumerate}
\vspace{-0.5\baselineskip}
where $N_{\rm inc} := \# \{ i = w_{j-1}, \cdots, w_j - 1 \mid f(\iter{\bs{x}}{i+1})>f(\iter{\bs{x}}{i}) \}$ and $\iter{f_{\max}}{k} := \max \{f(\iter{\bs{x}}{i}) \mid i = 1, \cdots, k \}$.

\vspace{-4.0\baselineskip}
\begin{center}
\begin{table*}
\vskip -0.1in

\caption{
The ASR of APGD and ACG for the robust models listed in RobustBench. CIFAR-100 was used as the dataset and $\varepsilon=\frac{8}{255}$.
The highest ASR is in bold, and the second is underlined. APGD($N$) refers to APGD with $N$-times the initial point selection. The meanings of the other columns are the same. 
}
\centering
\vskip 0.1in
\begin{small}
\begin{tabular}{ c | c | cccc || c}
\toprule
CIFAR-100 ($\varepsilon=8/255$) & & \multicolumn{4}{c}{Attack Success Rate}\\ \midrule
\bf{paper} & \bf{Architecture} & \bf{APGD(1)} & \bf{ACG(1)} & \bf{APGD(5)} & \bf{ACG(5)} & \bf{diff} \\ 
\hline\hline
\cite{Addepalli2021} & PreActResNet-18 & 72.10 & 72.12 & \underline{72.25} & \bf{72.47} & 0.22\\
\cite{Rade2021} & PreActResNet-18 & 70.40 & \underline{70.63} & 70.55 & \bf{70.86} & 0.31\\
\cite{Rebuffi2021} & PreActResNet-18 & 70.86 & \underline{71.07} & 70.93 & \bf{71.29} & 0.36\\
\cite{Rice2020} & PreActResNet-18 & 79.78 & \underline{80.24} & 79.99 & \bf{80.63} & 0.64\\
\cite{Hendrycks2019} & WideResNet-28-10 & 69.28 & \underline{69.97} & 69.50 & \bf{70.51} & 1.01\\
\cite{Rebuffi2021} & WideResNet-28-10 & 66.41 & \underline{66.87} & 66.67 & \bf{67.27} & 0.60\\
\cite{Addepalli2021} & WideResNet-34-10 & \underline{68.53} & 68.12 & \bf{68.74} & 68.52 & -0.21\\
\cite{Chen2021} & WideResNet-34-10 & 68.24 & \underline{68.42} & 68.36 & \bf{68.77} & 0.41\\
\cite{Cui2020} & WideResNet-34-10 & 71.85 & \underline{72.16} & 72.15 & \bf{72.56} & 0.41\\
\cite{Cui2020} & WideResNet-34-10 & 69.63 & \underline{69.96} & 69.87 & \bf{70.33} & 0.46\\
\cite{Sitawarin2020} & WideResNet-34-10 & 73.07 & \underline{73.64} & 73.43 & \bf{74.27} & 0.84\\
\cite{Wu2020a} & WideResNet-34-10 & 69.13 & \underline{69.58} & 69.32 & \bf{70.11} & 0.79\\
\cite{chen2020Efficient} & WideResNet-34-10 & 71.76 & 71.78 & \underline{71.96} & \bf{72.18} & 0.22\\
\cite{Cui2020} & WideResNet-34-20 & 68.50 & \underline{68.75} & 68.72 & \bf{69.13} & 0.41\\
\cite{Gowal2020} & WideResNet-70-16 & 61.23 & \underline{61.67} & 61.55 & \bf{62.19} & 0.64\\
\cite{Gowal2020} & WideResNet-70-16 & 68.76 & \underline{69.13} & 69.04 & \bf{69.43} & 0.39\\
\cite{Rebuffi2021} & WideResNet-70-16 & 63.94 & \underline{64.38} & 64.17 & \bf{64.77} & 0.60\\
\bottomrule
\end{tabular}
\end{small}
\vskip -0.1in
\label{table:cifar100_large}
\end{table*}
\end{center}

\begin{center}
\begin{table*}
\centering
\caption{Same as in Table~\ref{table:cifar100_large}, but using CIFAR-10 and ImageNet datasets with $\varepsilon=\frac{8}{255},\frac{4}{255}$, respectively.
}
\vskip 0.1in
\begin{small}
\begin{tabular}{ c | c | cccc || c}
    \toprule
    \centering
    CIFAR-10 ($\varepsilon=8/255$) & & \multicolumn{4}{c}{Attack Success Rate}\\ \midrule
    \bf{paper} & \bf{Architecture} & \bf{APGD(1)} & \bf{ACG(1)} & \bf{APGD(5)} & \bf{ACG(5)} & \bf{diff} \\ 
    \hline\hline
    \cite{Rade2021} & PreActResNet-18 & 42.33 & \underline{42.49} & 42.46 & \bf{42.65} & 0.19\\
    \cite{Rade2021} & PreActResNet-18 & 41.51 & \underline{41.84} & 41.65 & \bf{42.12} & 0.47\\
    \cite{Rebuffi2021} & PreActResNet-18 & 42.73 & \underline{43.01} & 42.91 & \bf{43.15} & 0.24\\
    \cite{andriushchenko2020square} & PreActResNet-18 & 53.55 & \underline{54.42} & 53.82 & \bf{54.90} & 1.08\\
    \cite{Sehwag2021} & ResNet-18 & 43.62 & \underline{44.16} & 43.91 & \bf{44.79} & 0.88\\
    \cite{Chen2020a} & ResNet-50 & 47.95 & \underline{48.12} & 48.08 & \bf{48.28} & 0.20\\
    \cite{Wong2020} & ResNet-50 & 54.02 & \underline{54.75} & 54.26 & \bf{55.44} & 1.18\\
    \cite{engstrom2019} & ResNet-50 & 47.69 & \underline{48.43} & 48.08 & \bf{49.25} & 1.17\\
    \cite{Rebuffi2021} & WideResNet-106-16 & 34.43 & 34.70 & \underline{34.71} & \bf{35.03} & 0.32\\
    \cite{Carmon2019} & WideResNet-28-10 & 39.38 & \underline{39.68} & 39.59 & \bf{40.03} & 0.44\\
    \cite{Gowal2020} & WideResNet-28-10 & 36.33 & \underline{36.63} & 36.45 & \bf{36.90} & 0.45\\
    \cite{Hendrycks2019} & WideResNet-28-10 & 43.53 & \underline{43.97} & 43.82 & \bf{44.36} & 0.54\\
    \cite{Rade2021} & WideResNet-28-10 & 38.48 & 38.62 & \underline{38.64} & \bf{38.87} & 0.23\\
    \cite{Rebuffi2021} & WideResNet-28-10 & 38.29 & 38.43 & \underline{38.47} & \bf{38.80} & 0.33\\
    \cite{Sehwag2020} & WideResNet-28-10 & 41.75 & \underline{42.07} & 41.93 & \bf{42.41} & 0.48\\
    \cite{Sridhar2021} & WideResNet-28-10 & 39.27 & \underline{39.49} & 39.45 & \bf{39.85} & 0.40\\
    \cite{Wang2019} & WideResNet-28-10 & 41.85 & 42.12 & \underline{42.15} & \bf{42.57} & 0.42\\
    \cite{Wu2020a} & WideResNet-28-10 & 39.38 & 39.49 & \underline{39.56} & \bf{39.70} & 0.14\\
    \cite{Zhang2020} & WideResNet-28-10 & 39.79 & 39.93 & \underline{39.98} & \bf{40.25} & 0.27\\
    \cite{Ding2019} & WideResNet-28-4 & 48.73 & \underline{53.40} & 49.67 & \bf{55.77} & 6.10\\
    \cite{Cui2020} & WideResNet-34-10 & 46.20 & \underline{46.42} & 46.41 & \bf{46.90} & 0.49\\
    \cite{Huang2020} & WideResNet-34-10 & 46.09 & \underline{46.30} & 46.19 & \bf{46.72} & 0.53\\
    \cite{Rade2021} & WideResNet-34-10 & 36.30 & \underline{36.57} & 36.46 & \bf{36.83} & 0.37\\
    \cite{Sehwag2021} & WideResNet-34-10 & 39.32 & \underline{39.84} & 39.58 & \bf{40.18} & 0.60\\
    \cite{Sitawarin2020} & WideResNet-34-10 & 46.91 & \underline{47.58} & 47.23 & \bf{48.02} & 0.79\\
    \cite{Wu2020a} & WideResNet-34-10 & 43.21 & 43.24 & \underline{43.36} & \bf{43.60} & 0.24\\
    \cite{Zhang2019a} & WideResNet-34-10 & 52.79 & \underline{53.46} & 53.08 & \bf{54.15} & 1.07\\
    \cite{Zhang2019} & WideResNet-34-10 & 46.44 & \underline{46.75} & 46.65 & \bf{47.18} & 0.53\\
    \cite{Zhang2020a} & WideResNet-34-10 & 45.54 & \underline{45.78} & 45.68 & \bf{46.12} & 0.44\\
    \cite{chen2020Efficient} & WideResNet-34-10 & 47.33 & 47.47 & \underline{47.58} & \bf{48.00} & 0.42\\
    \cite{Sridhar2021} & WideResNet-34-15 & 38.75 & \underline{38.93} & 38.90 & \bf{39.15} & 0.25\\
    \cite{Cui2020} & WideResNet-34-20 & 45.63 & \underline{45.91} & 45.88 & \bf{46.23} & 0.35\\
    \cite{Gowal2020} & WideResNet-34-20 & 42.55 & \underline{42.72} & 42.65 & \bf{42.86} & 0.21\\
    \cite{Pang2020} & WideResNet-34-20 & 44.54 & \underline{44.96} & 44.75 & \bf{45.33} & 0.58\\
    \cite{Rice2020} & WideResNet-34-20 & 44.69 & \underline{45.25} & 44.92 & \bf{45.69} & 0.77\\
    \cite{Huang2021} & WideResNet-34-R & 36.12 & \underline{36.34} & 36.27 & \bf{36.76} & 0.49\\
    \cite{Huang2021} & WideResNet-34-R & 37.10 & 37.27 & \underline{37.33} & \bf{37.79} & 0.46\\
    \cite{Gowal2020} & WideResNet-70-16 & 33.24 & \underline{33.49} & 33.42 & \bf{33.70} & 0.28\\
    \cite{Gowal2020} & WideResNet-70-16 & 41.95 & \underline{42.20} & 42.12 & \bf{42.45} & 0.33\\
    \cite{Gowal2021} & WideResNet-70-16 & 32.25 & \underline{32.71} & 32.57 & \bf{33.04} & 0.47\\
    \cite{Rebuffi2021} & WideResNet-70-16 & 32.28 & 32.44 & \underline{32.46} & \bf{32.75} & 0.29\\
    \cite{Rebuffi2021} & WideResNet-70-16 & 34.76 & 34.95 & \underline{35.04} & \bf{35.27} & 0.23\\

    \midrule
    ImageNet ($\varepsilon=4/255$) &\multicolumn{2}{c}{}&&\\ \hline\hline
\cite{DBLP:journals/corr/abs-2007-08489} & ResNet-18 & 72.80 & \underline{73.34} & 73.00 & \bf{73.72} & 0.72\\
\cite{DBLP:journals/corr/abs-2007-08489} & ResNet-50 & 62.72 & \underline{63.06} & 62.86 & \bf{63.70} & 0.84\\
\cite{Wong2020} & ResNet-50 & 71.58 & 71.64 & \underline{71.70} & \bf{71.94} & 0.24\\
\cite{engstrom2019} & ResNet-50 & 67.74 & \underline{68.08} & 67.86 & \bf{68.60} & 0.74\\
\cite{DBLP:journals/corr/abs-2007-08489} & WideResNet-50-2 & 58.88 & \underline{59.36} & 58.96 & \bf{59.92} & 0.96\\
\midrule
\multicolumn{7}{c}{\textbf{Summary}}\\
\midrule
\multicolumn{2}{c|}{the number of \textbf{bold models}} & 0 & 0 &  1 &  63 & \\
\multicolumn{2}{c|}{the number of \underline{underlined models}} & 1 & 49 & 14 & 0 &
\\
    \bottomrule
    \end{tabular}
    \end{small}
\vskip -0.1in
\label{table:imagenet_large}
\label{table:cifar10_large}
\end{table*}
\end{center}
\section{Experiments}
\label{sec:experiments}
We investigated the performance of ACG for an $l_\infty$ attack using the robust models listed in RobustBench.
\paragraph{Models and Dataset:}
We used 64 models, i.e., 42, 17, and 5 models based on the CIFAR-10, CIFAR-100, and ImageNet datasets, respectively. 
From a validation dataset, we used 10,000 test images for the evaluation when applying the CIFAR-10 and CIFAR-100 datasets, and 5,000 images when using the ImageNet dataset.
\paragraph{Loss Function:}
We used the CW loss \cite{carlini2017towards} as the objective function. Let $c$ be the correct answer class for input $\bs{x}$. 
Then, the CW loss is defined as follows.
\begin{align}
    \label{eq:cw_loss}
    {\rm CW}(\bs{x}, c) = -g_c(\bs{x}) + \maxop_{i\neq c} g_i(\bs{x}).
\end{align}
An attack using CW loss succeeds if we find an adversarial example $\xadv$ that satisfies ${\rm CW}(\xadv, c)\geq0$.
The results of experiments using the DLR loss proposed in \cite{croce2020reliable} are shown in  \Cref{results_of_dlr_loss}.
\paragraph{Initial Points.}
In the case of an $l_\infty$ attack, the center of the feasible region ($l_\infty$-ball with a diameter of $\varepsilon$) is defined as $ \frac{\bs{u}+\bs{\ell}}{2} $, where $\bs{u} = \min (\xorig + \varepsilon \bs{1}, \bs{1})$, and $\bs{\ell} = \max ( \xorig - \varepsilon \bs{1}, \bs{0})$. We referred to one hundred iterations of the update from an initial point as a restart. 
Following APGD, we applied 5 restarts in the experiments. 
The initial point of the first restart was the center of the feasible region, whereas the others were
determined through sampling from a uniform distribution.
For reproducibility of the results, the seed of the random numbers was fixed.

\subsection{Comparison of ACG and APGD}
To evaluate the attack performance under the formulation \Cref{eq:formulation_adex}, we conducted experiments to compare the performance of APGD, a SOTA method, with that of ACG. 
The parameters for the step size selection $\rho$, checkpoints $w$, the number of iterations $N_\mathrm{iter}$, and the number of restarts were the same as in the study on APGD, i.e., $\rho=0.75$, $N_{{\rm iter}}=100$, and 5 restarts.
The results are summarized in \Cref{table:cifar10_large,table:cifar100_large,table:imagenet_large}.
The columns APGD($N$) and ACG($N$) show the attack success rate (ASR) of APGD and ACG with $N$ restarts, respectively.
From these tables, it may be observed that the ASRs of ACG were higher than those of APGD in all but only 1 of the 64 models.
These results indicate that ACG exhibited a higher attack performance, regardless of the datasets or the model's architectures used.

Surprisingly, when comparing APGD(5) and ACG(1), it may be noted that ACG(1) achieved the same or higher ASR for three-fourths of all 64 models with fewer restarts (see \Cref{table:cifar10_large,table:cifar100_large,table:imagenet_large}). 
Therefore, we expect that ACG will enable faster attacks with fewer inferences. Details are provided in the next section.
Furthermore, ACG(1) does not rely on random numbers to select an initial point because the initial point is the center of the feasible region.
In other words, ACG outperforms APGD with only deterministic operations.

\begin{figure}[tb]
    \vskip 0.1in
    \centering
    \includegraphics[width=0.75\linewidth]{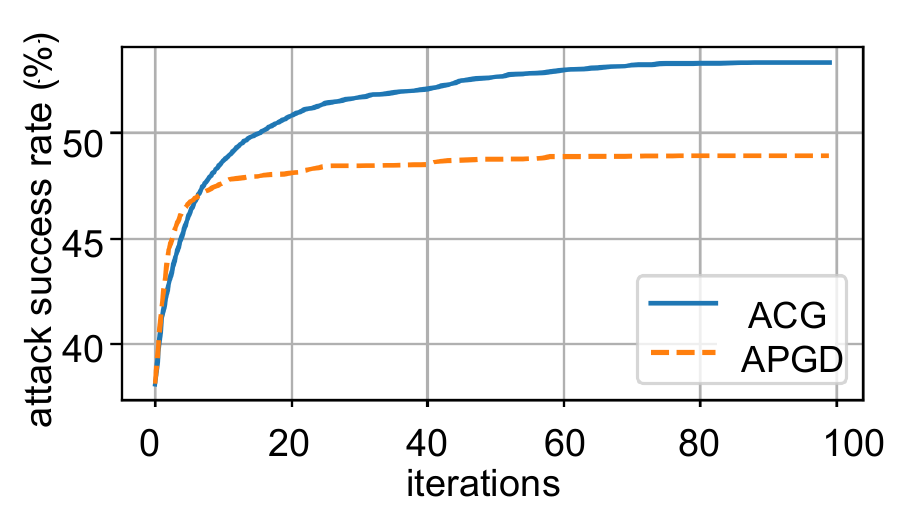}
    \caption{Transitions of the ASR of APGD and ACG.
    The attack model was \cite{Ding2019}.}
    \label{fig:best_loss_and_ASR}
    \vskip -0.1in
\end{figure}

\Cref{fig:best_loss_and_ASR} shows the transition of the ASR for APGD and ACG. The ASR of APGD rapidly increased and almost converged in the early stages of the search. By contrast, the ASR of ACG continues to increase even at the end of the search.
Note that this figure is created based on the data obtained from a model \cite{Ding2019}, whereas the same trend is also observed for other models.

\subsection{Comparison of execution time}
\label{execution_time}
We compared the execution times of APGD and ACG with Intel(R) Xeon(R) Gold 6240R CPU and NVIDIA GeForce RTX 3090 GPU. The execution time was recorded as the time elapsed from the start to the end of the attack. \Cref{table:executionTime} shows that APGD with five restarts (APGD(5)) took 22m, 5.88s to attack \cite{Ding2019}, and ACG with five restarts (ACG(5)) took 21m, 15.67s in real time. In addition, APGD with one restart (APGD(1)) took 6m, 45.26s to attack \cite{Ding2019}, and ACG with one restart (ACG(1)) took 6m, 56.78s. From this experiment, the ratio of the execution time of ACG(5) to APGD(5) was about 0.96 and that of ACG(1) to APGD(1) was approximately 1.03. This means the computational cost of ACG is nearly the same as that of APGD. Furthermore, ACG with one restart outperformed APGD with five restarts in terms of the ASR, which show that ACG was able to achieve a higher ASR more than 3 times faster than APGD.
\begin{small}
\begin{table*}[tb]
\centering
\caption{The average execution time of APGD and ACG. We measure the execution time 5 times with Intel(R) Xeon(R) Gold 6240R CPU @ 2.40GHz and NVIDIA GeForce RTX 3090 GPUs.}
\label{table:executionTime}
\vskip 0.15in
\begin{tabular}{c|cccc || c c c}
\toprule
\cite{Ding2019} & APGD(1) & ACG(1) & APGD(5) & ACG(5) & CPU & RAM & GPU\\
\midrule
ASR  &    48.73     &    53.40    &     49.67    &     55.77    &Intel(R) Xeon(R) & 786GB &NVIDIA\\
time &    6m45.26s     &    6m56.78s    &    22m5.88s     &   21m15.67s &Gold 6240R& &GeForce\\
ratio & 0.97 & 1 & 3.18 & 3.06 &$\times2$& &RTX 3090$\times 4$\\
\bottomrule
\end{tabular}
\vskip -0.1in
\end{table*}
\end{small}

\subsection{Variations in the Class Attacked}
\label{subsection:switchingOfCTC}
\begin{table}[t]
    \centering
    \caption{
    Average percentage of successful attacks and different CTC. 
    "Success(xx.xx\%)" indicates the percentage of images with successful attacks among all attacked images, and "Failure(xx.xx\%)" indicates that for images with failed attacks. The "Different CTC" column shows the percentage of the images in which ACG attacks have a different CTC from APGD in the corresponding row.
    }
    \vskip 0.1in
    \begin{small}
    \begin{tabular}{ c | c | c  }
    \toprule
    APGD & ACG & Different CTC\\
    \midrule
\multirow{2}{*}{Success(42.33\%)} & Success(42.27\%)& 6.18\% \\
& Failure(0.06\%)&11.96\%\\
\midrule
\multirow{2}{*}{Failure(57.67\%)} & Success(0.68\%)& \textbf{96.14}\%\\
& Failure(56.99\%)& 6.05\%\\
     \bottomrule
    \end{tabular}
    \end{small}
    \label{table:label_and_attack_success}
\end{table}

When we attacked using the CW loss, the search was updated to find inputs classified in class $c' \neq c$.
Specifically, $c'$ is $\argmax_{i \neq c} g_i(\bs{x})$, and we refer to this $c'$ as the \emph{CW target class} (\emph{CTC}).
The value of $c'$ varied during the search, depending on the search point.
This section summarizes the CTC results of ACG and APGD attacks based on the results of the CIFAR-10 dataset.

First, CTCs were changed at least once during the search in 42.47\% of inputs for ACG and 1.36\% for APGD.
The average number of times the CTC was switched during the search was 2.14 for ACG and 0.02 for APGD.
In APGD, CTC showed almost no change from the initial CTC during the search.
In contrast, ACG frequently switched CTCs during the search.
\Cref{table:label_and_attack_success} also summarizes the differences in the final CTCs of ACG and APGD in terms of successful and unsuccessful attacks.
As shown in the table, ACG attacked a different class than APGD for 96.14\% of the images that failed in APGD but succeeded in ACG.
The same trend was observed in the other models, suggesting that ACG increased the ASR by switching CTCs and attacking a different class than APGD.

\subsection{Effect of Conjugate Gradient}
\label{subsection:effectOfConjugate}
\begin{figure}[tb]
    \centering
    \includegraphics[width=0.9\linewidth]{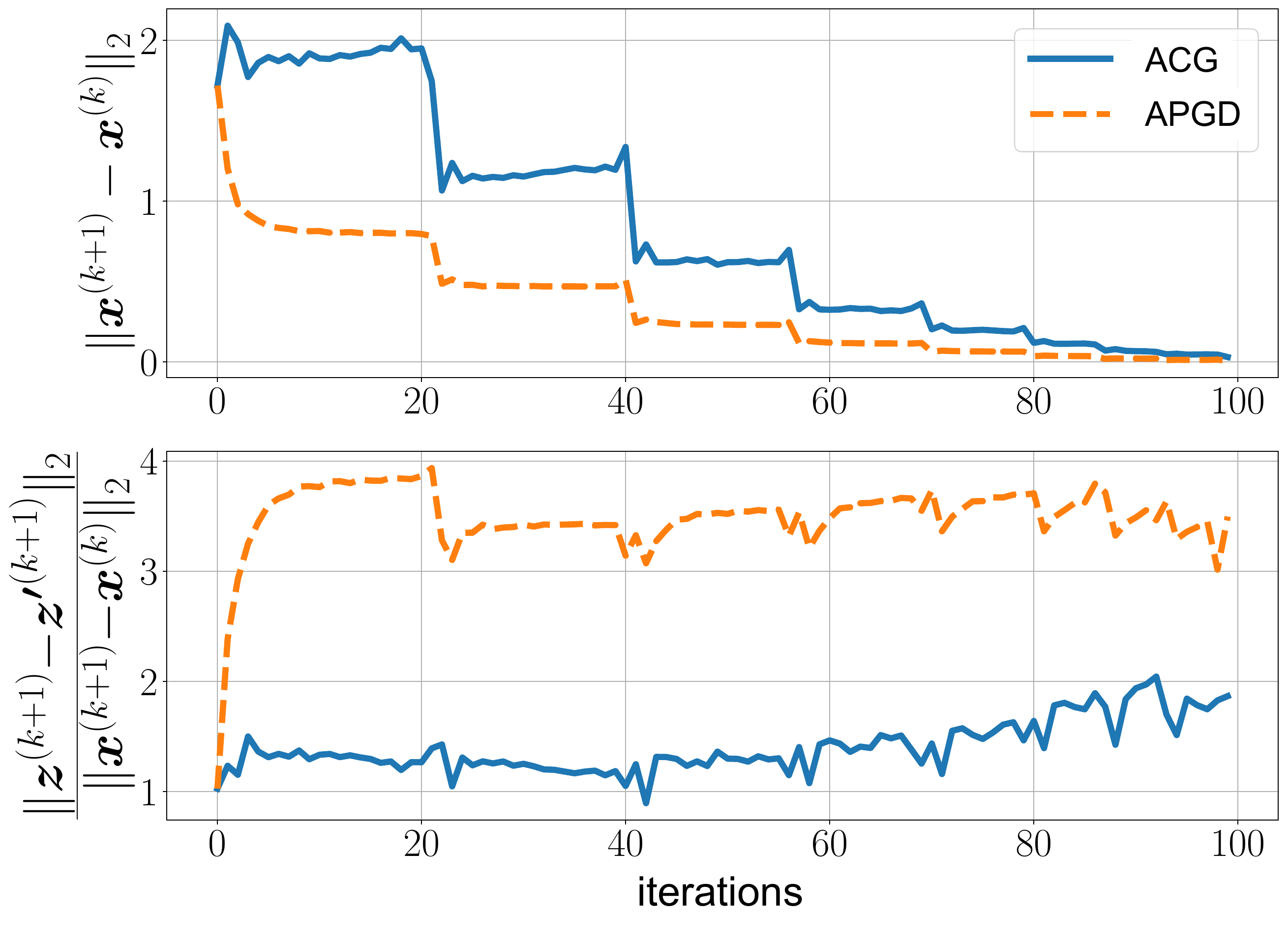}
    \caption{
    Effects of the projection on the feasible region and moving distances of APGD and ACG. The upper part shows the 2-norm of two successive points. The lower part shows the ratio of the 2-norm of the two consecutive points to the 2-norm of the projection onto the feasible region.
    Of note, $\iter{\bs{z}}{k+1}=\iter{\bs{x}}{k+1}$ for ACG.
    }
    \label{fig:move_distance}
\end{figure}

\Cref{table:cifar10_large,table:cifar100_large} shows that ACG achieved a higher ASR value than APGD.
Because the update function is the major difference between the two algorithms (see \Cref{table:differenceAcgGcgAPGD}), we investigated how the conjugate gradient affected the search (see the formula \Cref{eq:update_sk}).

The top portion of ~\Cref{fig:move_distance} shows the transitions of the 2-norm between two successive search points $\|\iter{\bs{x}}{k+1} - \iter{\bs{x}}{k} \|_2$ on APGD and ACG. Regarding the distance moved between the two points, it may be observed that the search points of ACG moved further than those of APGD.  Moreover, to investigate the effect of the projection on APGD, we calculated $\frac{\| \iter{\bs{z}}{k+1} - \iter{\bs{z'}}{k+1} \|_2}{\|\iter{\bs{x}}{k+1} - \iter{\bs{x}}{k}\|_2}$ as the ratio of the distance traveled between the two search points, which indicates the amount of update distance wasted by the projection. 
Although we mainly focused on the numerator $\| \iter{\bs{z}}{k+1} - \iter{\bs{z'}}{k+1} \|_2$, we divided it by $\|\iter{\bs{x}}{k+1} - \iter{\bs{x}}{k}\|_2$ to exclude the effect of the difference in step size.
The ratio of the distance traveled in 100 iterations is shown at the bottom of \Cref{fig:move_distance}.
APGD exhibited a higher ratio of projection in the distance traveled than ACG.
That is, in the update of APGD after moving $\nabla f(\iter{\bs{x}}{k})$, the projection returns it to the vicinity of the original point $\iter{\bs{x}}{k-1}$. Therefore, we can see that ACG moves more than APGD owing to the introduction of the conjugate direction.

%

\section{Search Diversity Analysis of ACG}
\label{section:studyOfACG}
The experimental results described in Section \ref{sec:experiments} show that ACG exhibited a higher ASR than APGD.
In addition,
the Euclidean distance between the successive two points of ACG was larger than that of APGD (see \Cref{subsection:effectOfConjugate}), which suggests that the search of ACG was more diverse. 
In this section, we verify this hypothesis.
While a search is intensified when the distance between successive two points is small (see the white points in \Cref{fig:DiversificationAndIntensification}), the search is not always diversified even if the successive two points are distant (see bottom of \Cref{example_of_diversity_index}).
We then regard the search as intensified (diversified) when the search points (do not) form clusters and propose an index that measures the degree of diversification by utilizing a global clustering coefficient.



\subsection{Definition of \measurementFirst}
\label{subsection:measurement}
The \emph{global clustering coefficient} \cite{Kemper2010} represents the strength of the connections between the nodes of a graph, and is often used in complex network analysis \cite{Tabak2014,Said2018}.
To apply the global clustering coefficient to our analysis, we consider a graph whose nodes are search points.

Given the set of search points $X$,
we define a graph $G_{X}(\theta) := (X, E(\theta))$, where
$E(\theta)=\{(\bs{v}, \bs{w}) \in X \times X \mid \|\bs{v}-\bs{w}\|_2 \leq \theta\}$.
Let $C(G)$ be the global clustering coefficient of a graph $G$, and let $h(\theta; X)$ be $1 - C(G_X(\theta))$. Note that $0 \le C(G_X(\theta)) \le 1$ from the definition of the global clustering coefficient (see \Cref{appendix:clustering_coefficient}).
Of note, when $G_X(\theta)$ is disjoint union of some complete graphs, $C(G_X(\theta)) = 1$ and thus $h(\theta; X) = 0$. Similarly, $h(\theta; X)$ takes low value when $G_X(\theta)$ exhibits clusters.
When the cluster structure of the search point is apparent, the cluster structure appears in $G_X(\theta)$ even for a small $\theta$. 
Therefore, we can quantify the cluster structure of the search points by the transition of $h$ with the change of $\theta$.

Hence, we define the \emph{\measurementFirst}(\measurementNoSpace) as the average of $h(\theta;X)$ for $\theta$ to quantify the diversity of the search points, as given below.
\begin{align*}
    \mathrm{\measurementNoSpace}(X, M) &:= \frac{1}{M}\int_{0}^{M} h(\theta; X)~d\theta,
\end{align*}
where $M = \sup \{\|\bs{x}-\bs{y}\|_2 \mid \bs{x}, \bs{y} \in \mathcal{S}\}$ is the size of the feasible region.
Because we consider an adversarial attack with an $l_\infty$-ball constraint, we obtain $M = \| \bs{u} - \bs{\ell} \|_2$.
When $X$ and $M$ are obvious, we simply denote $\mathrm{\measurementNoSpace}(X, M)$ as \measurementNoSpace. 
From the definition of \measurementNoSpace, we obtain $0 \leq \mathrm{\measurementNoSpace} \leq 1$.

\begin{figure}[tb]
    \centering
    \includegraphics[width=0.95\linewidth]{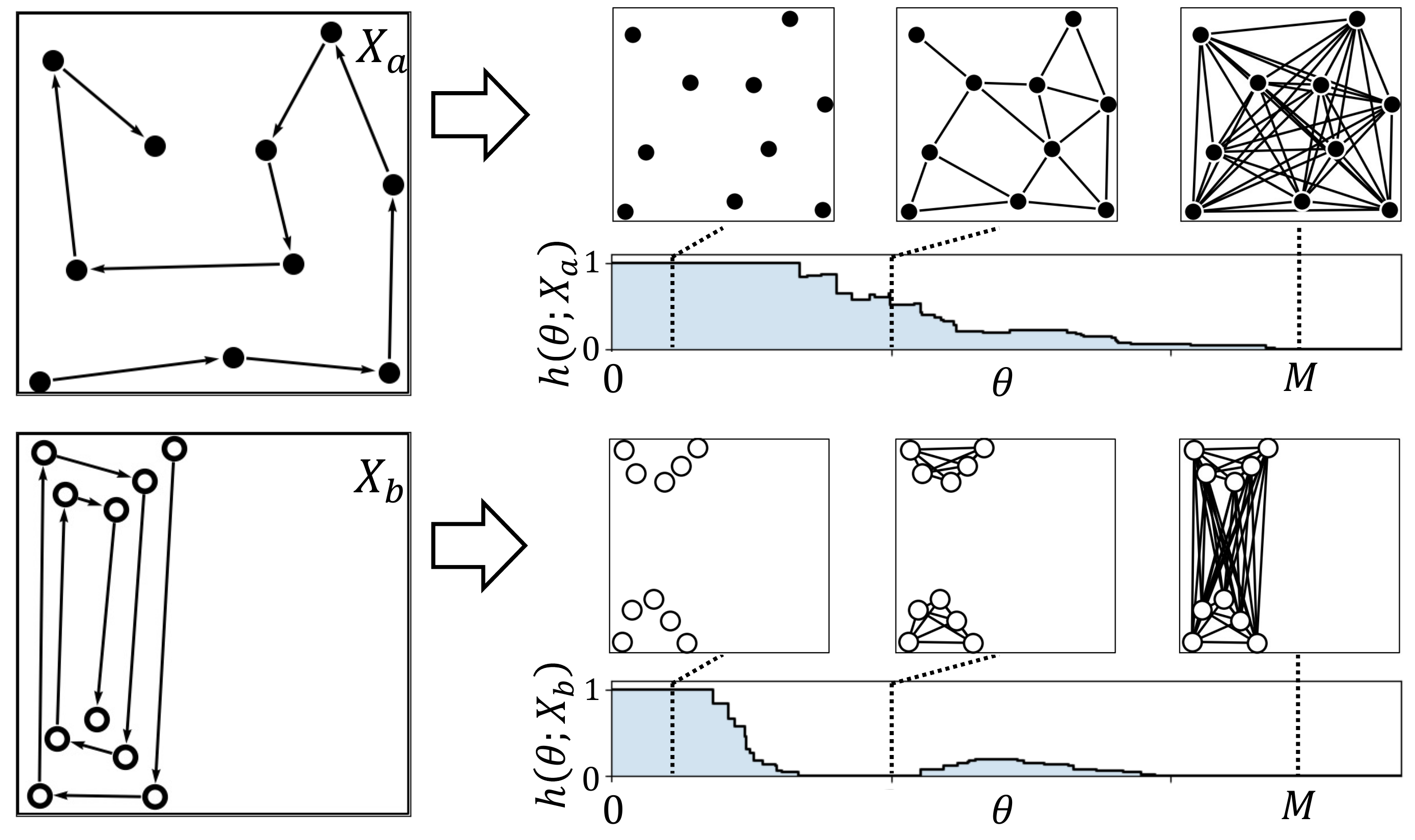}
    \caption{Example of the relationship between the diversity of search points and the global clustering coefficients. The white and black circles are the same as in \Cref{fig:DiversificationAndIntensification}.
    The left side shows the movements of the search points $X$.
    The right side shows the transitions of $h(\theta; X)$, and three graphs for several thresholds $\theta$.
    }
    \label{example_of_diversity_index}
    \vskip -0.2in
\end{figure}

In \Cref{example_of_diversity_index}, examples of $h(\theta; X)$ are shown for two sets of search points, $X_a$ (upper) and $X_b$ (lower). 
Because $X_b$ includes two clusters, whereas $X_a$ does not, we regard $X_a$ as an example of a diverse search and $X_b$ as an intense search.
On the right side of \Cref{example_of_diversity_index}, it may be observed that $h(\theta; X_a) \ge h(\theta; X_b)$ for most $\theta \in [0,M]$ and thus 
$\mathrm{\measurementNoSpace}(X_a, M) > \mathrm{\measurementNoSpace}(X_b, M)$, which reflects the diversity of the search points.
In other words, $G_{X_b}(\theta)$ was more clustered than $G_{X_a}(\theta)$ for most  $\theta$.

As shown in \Cref{example_of_diversity_index}, the following relationship holds between \measurement and the diversity of the search points: when \measurement is small (large), the points in $X$ (do not) form some clusters.
That is, an intensive (diverse) search is conducted.
Below, we use \measurement to analyze and discuss APGD and ACG.

\begin{figure*}[t]
    \centering
    \begin{tabular}{ccc}
        \centering
        \includegraphics[valign=m,width=0.313\linewidth]{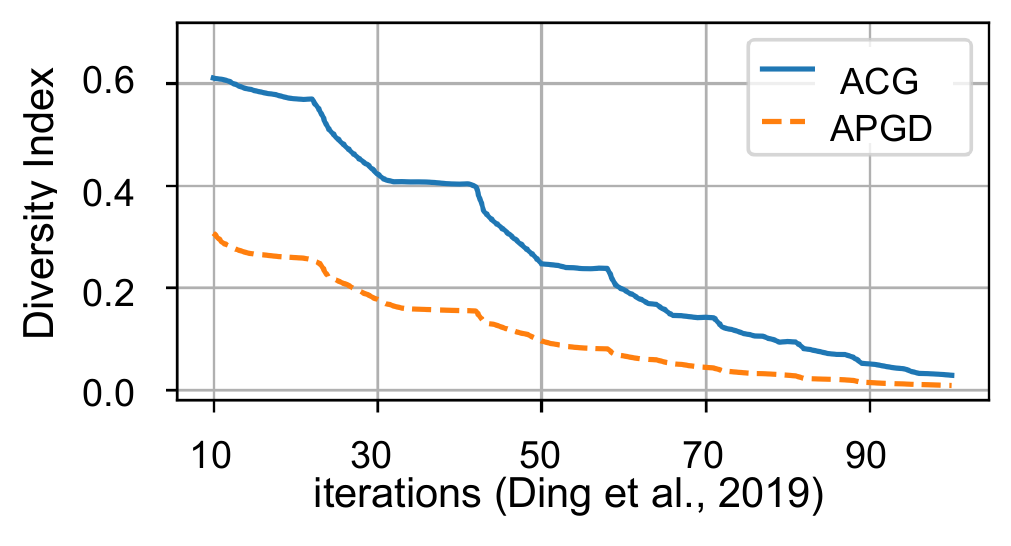}
      &
        \centering
        \includegraphics[valign=m,width=0.313\linewidth]{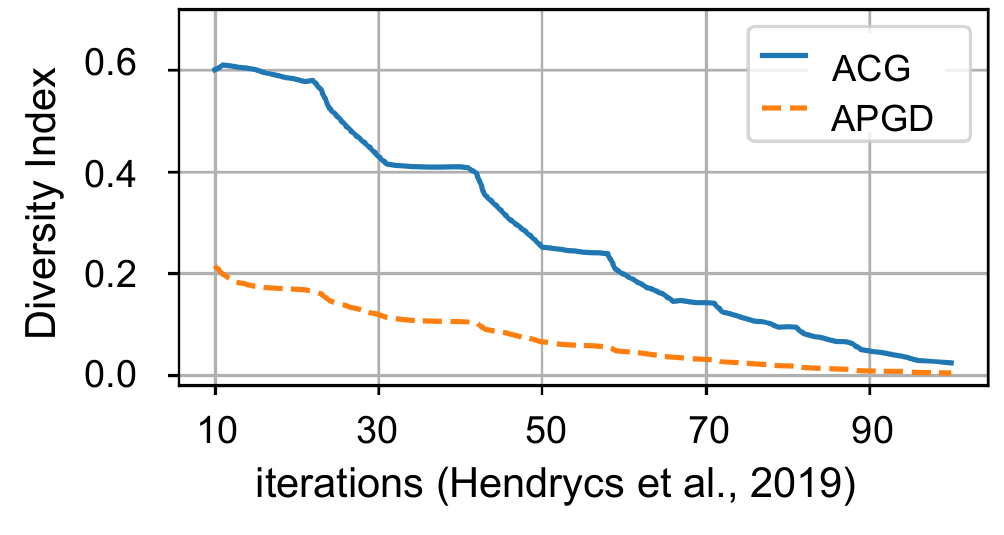}
      &
        \centering
        \includegraphics[valign=m,width=0.313\linewidth]{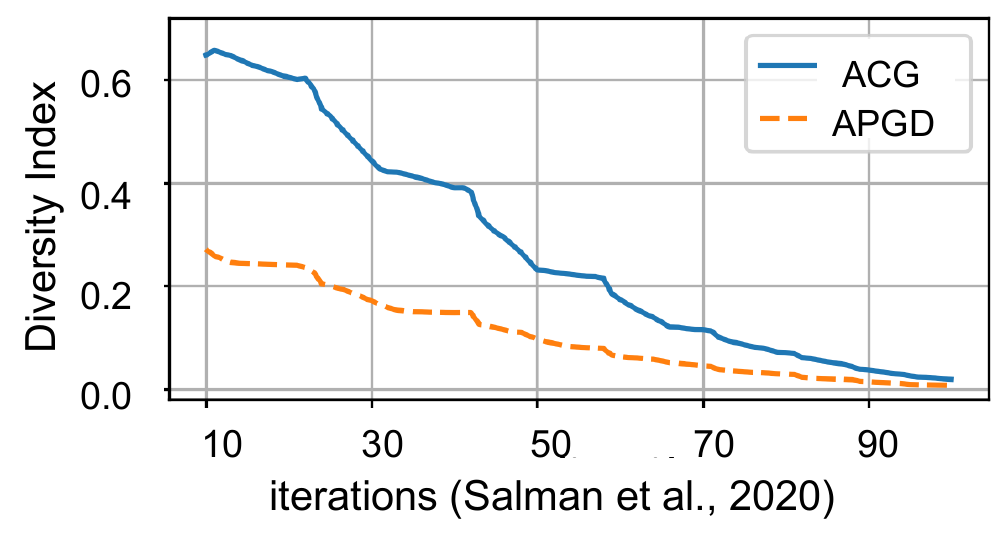}
    \end{tabular}
   \caption{Transition of the average \measurement over 10,000 images using three models.}
   \label{clustering_coef_APGD_and_ACG}
   \vskip 0.2in
   \includegraphics[width=0.8\linewidth]{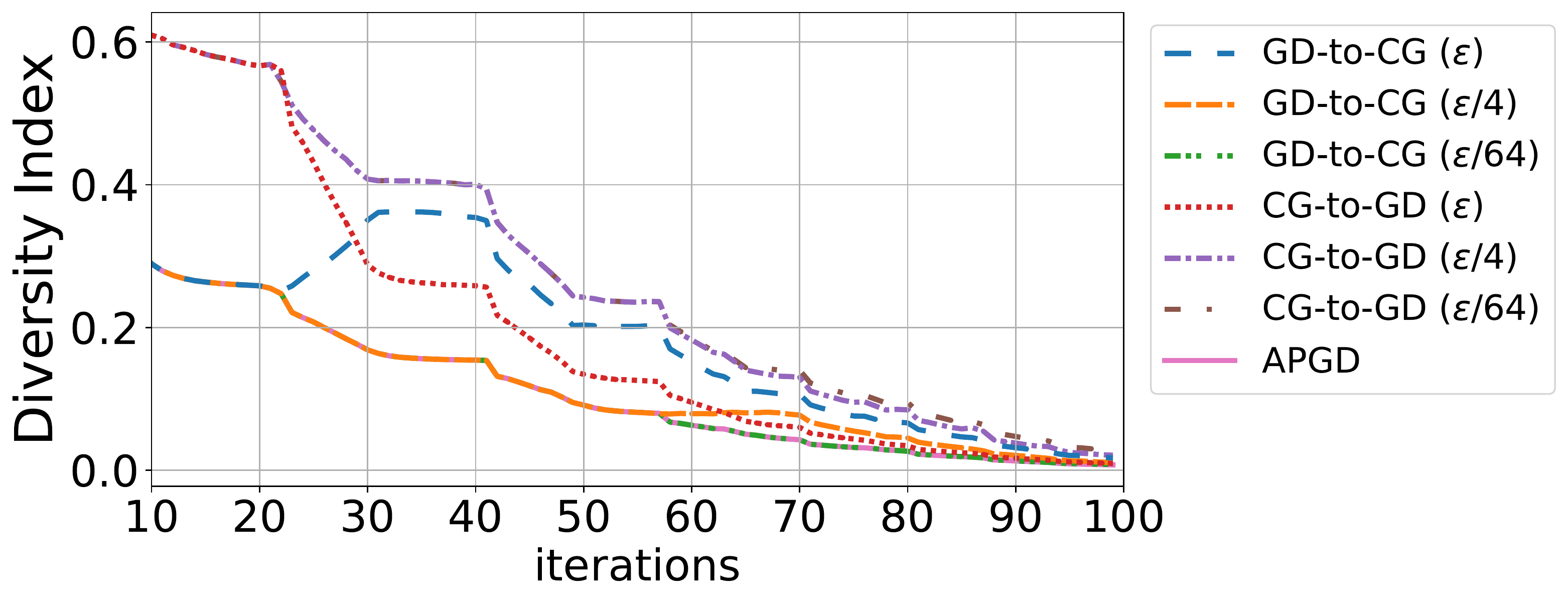}
   \caption{The transitions of {\measurementNoSpace} of the switching update direction from GD and CG averaged over 10,000 images.}
   \label{fig:ConjugatePGD_bestloss_and_DI}
\end{figure*}

\begin{table*}[tb]
\centering
\caption{
ASR when switching from GD to CG (GD-to-CG) or CG to GD (CG-to-GD) at the specific step size.
}
\vskip 0.1in
\begin{tabular}{cccccccc || cc}
\toprule
        step size & $\varepsilon$ & $\varepsilon/2$     & $\varepsilon / 4$   & $\varepsilon/8$     & $\varepsilon/16$     & $\varepsilon/32$     & $\varepsilon/64$     & APGD  & ACG   \\
\midrule
GD-to-CG &  50.85  & 49.70 & 49.66  & 49.66   & 49.67   & 49.67 & 49.67 & \multirow{2}{*}{49.67}  & \multirow{2}{*}{55.77} \\
CG-to-GD & 55.77  & 55.84 & 55.80   & 55.80   & 55.78   & 55.77   & 55.77  &  &  \\
\bottomrule
\end{tabular}
\label{table:PGDConjugate_ASR_at_specificstepsize}
\vskip -0.1in
\end{table*}

\subsection{Comparison on the behavior of \measurement}
\label{subsection:ComparisonOfApgdAndAcgAboutDI}
We considered the difference between the conjugate gradient and the gradient in the momentum direction using \measurementNoSpace.
In the following analyses, we use 
$\iter{X}{k} := \left\{\iter{\bs{x}}{k-9}, \cdots,  \iter{\bs{x}}{k} \right\}$ as a set of search points at $k$-th iteration where $\iter{\bs{x}}{i}$ is the $i$-th search point.
The diversity of the search and the transition of the search trends may be observed from \measurement calculated for the latest 10 search points.
\Cref{clustering_coef_APGD_and_ACG} shows the transitions of \measurement in each algorithm for the models whose results exhibited the most significant differences in the ASR for each dataset.
From \Cref{clustering_coef_APGD_and_ACG}, it may be observed that
the {\measurementNoSpace}s during the search for ACG were larger than those obtained in the search by APGD.
This suggests that ACG moves to a wide variety of different points and conducts more diverse searches. In particular, when the step size is large, ACG extensively explores the feasible region.
By contrast, APGD exhibited a small \measurement value from the early iterations during the search.
\Cref{subsection:effectOfConjugate} shows that the movement distance of the search point accompanied with an update of APGD was shorter than that of ACG. In addition, the results in this section show that the movement distance in the latest 10 search points of APGD was limited in comparison to that of ACG.

These results demonstrate that ACG conducted a more diverse search than APGD, which is one possible reason for the difference of CTC and the higher ASR of ACG.
We believe that utilizing \measurement to control the balance of diversification and intensification may be considered a promising approach to further improve ASR.

\subsection{Comparison of APGD and ACG on the ability to diversification and intensification}
\label{subsection:comparisonOfApgdAndAcgAboutDivesity}

In the previous section, we observed that ACG conducts a more diverse search than APGD.
In this section, we investigate how the diverse search for ACG contributes to the recorded increase in ASR.
We used the CIFAR-10 as the dataset and a WideResNet-28-4 model trained using the method proposed in \cite{Ding2019}, and CW loss as the objective function.
APGD utilizes the gradient direction (GD) as the update direction.
ACG and APGD utilize the same step-size selection strategy, which halves the step size from $2\varepsilon$ when either condition \eqref{eq:condition_1} or \eqref{eq:condition_2} is satisfied.

We use GD-to-CG ($\eta$) and CG-to-GD ($\eta$) to indicate that the update direction was changed at a step size of $\eta$ from GD to CG and CG to GD, respectively.
For example, GD-to-CG($\eta/4$) searches in GD when $\iter{\eta}{k}\geq \varepsilon/2$, and in CG when $\iter{\eta}{k}\leq\varepsilon/4 $.
In \Cref{fig:ConjugatePGD_bestloss_and_DI}, we observe that DI initially increased when switching from GD to CG, but did not increase when switching toward the end of the search.
In addition, we found that the DI decreased when switching from CG to GD, indicating that the diversification decreased.
By contrast, DI did not decrease in the latter half of the search, indicating that diversification did not significantly decrease.

\Cref{table:PGDConjugate_ASR_at_specificstepsize} shows the ASR results for 100 iterations and 5 restarts when switching from GD to CG or CG to GD at each step size. 
It may be observed from the table entry of CG-to-GD, we can see that the ASR when using CG in the early stage was higher than that of ACG.
In summary, the early stage of ACG performs a more diversified search than that of APGD, and the diversification in the early phase of ACG contributed to the higher ASR of ACG in comparison to that of APGD.
\section{Conclusion}
In this study, we have proposed ACG using conjugate directions inspired by the CG method.
We have conducted extensive experiments to evaluate the performance of the proposed approach on a total of 64 models listed in RobustBench. 
We found that ACG significantly improved the ASR compared to APGD in all but one of the models.
In particular, ACG with one restart outperformed the ASR of APGD with 5 restarts for many of the models. 
This indicates that ACG may be promising in that it achieves a high ASR with only deterministic operations. 
In addition, ACG frequently switched the attacked class during the search and succeeded in attacking owing to its attacked class being different from APGD, whereas APGD rarely switched the attacked class.
This result implies that varying the attacked class contributed significantly to the improvement in terms of ASR exhibited by ACG.

To analyze the difference in the search performance between ACG and APGD, we have proposed the \measurementFirst (\measurementNoSpace), which measures the degree of diversification of a search.
\measurement is calculated based on the global clustering coefficients on a graph the nodes of which are the latest among several search points. 
A higher \measurement indicates that corresponding search points are sparsely distributed.
According to the analyses using \measurementNoSpace, one of the reasons for the higher ASR of ACG is that the CG direction exhibited greater diversification, particularly in the early phase of the search.
\measurement may also be considered a valuable metric in analyzing other attack algorithms.
In the future, we expect that an algorithm designed to efficiently control diversification and intensification using \measurement would be effective.
\section*{Acknowledgement}
This research project was supported by the Japan Science and Technology Agency (JST), the Core Research of Evolutionary Science and Technology (CREST), the Center of Innovation Science and Technology based Radical Innovation and Entrepreneurship Program (COI Program), JSPS KAKENHI Grant Number JP20H04.
\clearpage
\bibliography{references.bib}

\begin{thebibliography}{52}
\providecommand{\natexlab}[1]{#1}
\providecommand{\url}[1]{\texttt{#1}}
\expandafter\ifx\csname urlstyle\endcsname\relax
  \providecommand{\doi}[1]{doi: #1}\else
  \providecommand{\doi}{doi: \begingroup \urlstyle{rm}\Url}\fi

\bibitem[Addepalli et~al.(2021)Addepalli, Jain, Sriramanan, Khare, and
  Babu]{Addepalli2021}
Addepalli, S., Jain, S., Sriramanan, G., Khare, S., and Babu, R.~V.
\newblock Towards achieving adversarial robustness beyond perceptual limits.
\newblock \emph{ICML 2021 Workshop}, 2021.

\bibitem[Andriushchenko et~al.(2020)Andriushchenko, Croce, Flammarion, and
  Hein]{andriushchenko2020square}
Andriushchenko, M., Croce, F., Flammarion, N., and Hein, M.
\newblock Square attack: a query-efficient black-box adversarial attack via
  random search.
\newblock In \emph{European Conference on Computer Vision}, pp.\  484--501.
  Springer, 2020.

\bibitem[Carlini \& Wagner(2017)Carlini and Wagner]{carlini2017towards}
Carlini, N. and Wagner, D.
\newblock Towards evaluating the robustness of neural networks.
\newblock In \emph{2017 ieee symposium on security and privacy (sp)}, pp.\
  39--57, 2017.

\bibitem[Carmon et~al.(2019)Carmon, Raghunathan, Schmidt, Liang, and
  Duchi]{Carmon2019}
Carmon, Y., Raghunathan, A., Schmidt, L., Liang, P., and Duchi, J.~C.
\newblock {Unlabeled Data Improves Adversarial Robustness}.
\newblock \emph{Advances in Neural Information Processing Systems}, 32, may
  2019.
\newblock ISSN 10495258.

\bibitem[Chen \& Lee(2021)Chen and Lee]{Chen2021}
Chen, E.-C. and Lee, C.-R.
\newblock Ltd: Low temperature distillation for robust adversarial training.
\newblock \emph{CoRR}, abs/2111.02331, 11 2021.

\bibitem[Chen et~al.(2021)Chen, Cheng, Gan, Gu, and Liu]{chen2020Efficient}
Chen, J., Cheng, Y., Gan, Z., Gu, Q., and Liu, J.
\newblock Efficient robust training via backward smoothing, 2021.

\bibitem[Chen et~al.(2020)Chen, Liu, Chang, Cheng, Amini, and Wang]{Chen2020a}
Chen, T., Liu, S., Chang, S., Cheng, Y., Amini, L., and Wang, Z.
\newblock {Adversarial Robustness: From Self-Supervised Pre-Training to
  Fine-Tuning}.
\newblock \emph{Proceedings of the IEEE Computer Society Conference on Computer
  Vision and Pattern Recognition}, pp.\  696--705, mar 2020.
\newblock ISSN 10636919.
\newblock \doi{10.1109/CVPR42600.2020.00078}.

\bibitem[Cheng et~al.(2014)Cheng, Shi, Qin, Zhang, and Bai]{Cheng2014}
Cheng, S., Shi, Y., Qin, Q., Zhang, Q., and Bai, R.
\newblock {Population Diversity Maintenance In Brain Storm Optimization
  Algorithm}.
\newblock \emph{Journal of Artificial Intelligence and Soft Computing
  Research}, 4\penalty0 (2):\penalty0 83--97, 2014.
\newblock ISSN 2083-2567.
\newblock \doi{10.1515/jaiscr-2015-0001}.

\bibitem[{\v{C}}repin{\v{s}}ekMatej et~al.(2013){\v{C}}repin{\v{s}}ekMatej,
  LiuShih-Hsi, and MernikMarjan]{CrepinsekMatej2013}
{\v{C}}repin{\v{s}}ekMatej, LiuShih-Hsi, and MernikMarjan.
\newblock {Exploration and exploitation in evolutionary algorithms}.
\newblock \emph{ACM Computing Surveys (CSUR)}, 45\penalty0 (3):\penalty0 33,
  jul 2013.
\newblock \doi{10.1145/2480741.2480752}.

\bibitem[Croce \& Hein(2020{\natexlab{a}})Croce and Hein]{croce2020minimally}
Croce, F. and Hein, M.
\newblock Minimally distorted adversarial examples with a fast adaptive
  boundary attack.
\newblock In \emph{International Conference on Machine Learning}, pp.\
  2196--2205. PMLR, 2020{\natexlab{a}}.

\bibitem[Croce \& Hein(2020{\natexlab{b}})Croce and Hein]{croce2020reliable}
Croce, F. and Hein, M.
\newblock Reliable evaluation of adversarial robustness with an ensemble of
  diverse parameter-free attacks.
\newblock In \emph{International Conference on Machine Learning}, pp.\
  2206--2216. PMLR, 2020{\natexlab{b}}.

\bibitem[Croce et~al.(2021)Croce, Andriushchenko, Sehwag, Debenedetti,
  Flammarion, Chiang, Mittal, and Hein]{croce2020robustbench}
Croce, F., Andriushchenko, M., Sehwag, V., Debenedetti, E., Flammarion, N.,
  Chiang, M., Mittal, P., and Hein, M.
\newblock Robustbench: a standardized adversarial robustness benchmark.
\newblock In \emph{Thirty-fifth Conference on Neural Information Processing
  Systems Datasets and Benchmarks Track}, 2021.

\bibitem[Cui et~al.(2021)Cui, Liu, Wang, and Jia]{Cui2020}
Cui, J., Liu, S., Wang, L., and Jia, J.
\newblock Learnable boundary guided adversarial training.
\newblock In \emph{Proceedings of the IEEE/CVF International Conference on
  Computer Vision (ICCV)}, pp.\  15721--15730, October 2021.

\bibitem[Ding et~al.(2020)Ding, Sharma, Lui, and Huang]{Ding2019}
Ding, G.~W., Sharma, Y., Lui, K. Y.~C., and Huang, R.
\newblock {MMA} training: Direct input space margin maximization through
  adversarial training.
\newblock In \emph{8th International Conference on Learning Representations,
  {ICLR} 2020, Addis Ababa, Ethiopia, April 26-30, 2020}. OpenReview.net, 2020.

\bibitem[Dong et~al.(2018)Dong, Liao, Pang, Su, Zhu, Hu, and
  Li]{dong2018boosting}
Dong, Y., Liao, F., Pang, T., Su, H., Zhu, J., Hu, X., and Li, J.
\newblock Boosting adversarial attacks with momentum.
\newblock In \emph{Proceedings of the IEEE conference on computer vision and
  pattern recognition}, pp.\  9185--9193, 2018.

\bibitem[Engstrom et~al.(2019)Engstrom, Ilyas, Salman, Santurkar, and
  Tsipras]{engstrom2019}
Engstrom, L., Ilyas, A., Salman, H., Santurkar, S., and Tsipras, D.
\newblock Robustness (python library), 2019.

\bibitem[Goodfellow et~al.(2015)Goodfellow, Shlens, and
  Szegedy]{goodfellow2014explaining}
Goodfellow, I.~J., Shlens, J., and Szegedy, C.
\newblock Explaining and harnessing adversarial examples.
\newblock In \emph{3rd International Conference on Learning Representations,
  {ICLR} 2015, San Diego, CA, USA, May 7-9, 2015, Conference Track
  Proceedings}, 2015.

\bibitem[Gowal et~al.(2019)Gowal, Uesato, Qin, Huang, Mann, and
  Kohli]{DBLP:journals/corr/abs-1910-09338}
Gowal, S., Uesato, J., Qin, C., Huang, P., Mann, T.~A., and Kohli, P.
\newblock An alternative surrogate loss for pgd-based adversarial testing.
\newblock \emph{CoRR}, abs/1910.09338, 2019.
\newblock URL \url{http://arxiv.org/abs/1910.09338}.

\bibitem[Gowal et~al.(2020)Gowal, Qin, Uesato, Mann, and Kohli]{Gowal2020}
Gowal, S., Qin, C., Uesato, J., Mann, T., and Kohli, P.
\newblock {Uncovering the Limits of Adversarial Training against Norm-Bounded
  Adversarial Examples}.
\newblock \emph{CoRR}, abs/2010.03593, oct 2020.

\bibitem[Gowal et~al.(2021)Gowal, Rebuffi, Wiles, Stimberg, Calian, and
  Mann]{Gowal2021}
Gowal, S., Rebuffi, S.-A., Wiles, O., Stimberg, F., Calian, D.~A., and Mann, T.
\newblock Improving robustness using generated data.
\newblock In \emph{NeurIPS}, 2021.

\bibitem[Hager \& Zhang(2006)Hager and Zhang]{Hager2006}
Hager, W. W.~W. and Zhang, H.
\newblock {A Survey of Nonlinear Conjugate Gradient Methods}.
\newblock \emph{Pacific journal of Optimization}, 2\penalty0 (1):\penalty0
  35--58, 2006.

\bibitem[Hendrycks et~al.(2019)Hendrycks, Lee, and Mazeika]{Hendrycks2019}
Hendrycks, D., Lee, K., and Mazeika, M.
\newblock Using pre-training can improve model robustness and uncertainty.
\newblock \emph{Proceedings of the International Conference on Machine
  Learning}, pp.\  2712--2721, 5 2019.
\newblock ISSN 2640-3498.

\bibitem[Hestenes \& Stiefel(1952)Hestenes and Stiefel]{Hestenes1952}
Hestenes, M. and Stiefel, E.
\newblock {Methods of conjugate gradients for solving linear systems}.
\newblock \emph{Journal of Research of the National Bureau of Standards},
  49\penalty0 (6):\penalty0 409, 1952.
\newblock ISSN 0091-0635.
\newblock \doi{10.6028/jres.049.044}.

\bibitem[Huang et~al.(2021)Huang, Wang, Erfani, Gu, Bailey, and Ma]{Huang2021}
Huang, H., Wang, Y., Erfani, S.~M., Gu, Q., Bailey, J., and Ma, X.
\newblock Exploring architectural ingredients of adversarially robust deep
  neural networks.
\newblock In \emph{NeurIPS}, 2021.

\bibitem[Huang et~al.(2020)Huang, Zhang, and Zhang]{Huang2020}
Huang, L., Zhang, C., and Zhang, H.
\newblock {Self-Adaptive Training: beyond Empirical Risk Minimization}.
\newblock \emph{Advances in Neural Information Processing Systems},
  2020-December, feb 2020.
\newblock ISSN 10495258.

\bibitem[Kemper(2010)]{Kemper2010}
Kemper, A.
\newblock {Valuation of Network Effects in Software Markets}.
\newblock 2010.
\newblock \doi{10.1007/978-3-7908-2367-7}.

\bibitem[Kurakin et~al.(2017)Kurakin, Goodfellow, and Bengio]{I-FGSM}
Kurakin, A., Goodfellow, I., and Bengio, S.
\newblock Adversarial examples in the physical world.
\newblock \emph{ICLR Workshop}, 2017.

\bibitem[Lin et~al.(2020)Lin, Song, He, Wang, and Hopcroft]{Lin2020Nesterov}
Lin, J., Song, C., He, K., Wang, L., and Hopcroft, J.~E.
\newblock Nesterov accelerated gradient and scale invariance for adversarial
  attacks.
\newblock In \emph{International Conference on Learning Representations}, 2020.

\bibitem[Madry et~al.(2018)Madry, Makelov, Schmidt, Tsipras, and
  Vladu]{madry2018towards}
Madry, A., Makelov, A., Schmidt, L., Tsipras, D., and Vladu, A.
\newblock Towards deep learning models resistant to adversarial attacks.
\newblock In \emph{International Conference on Learning Representations}, 2018.

\bibitem[Morales-Casta{\~{n}}eda et~al.(2020)Morales-Casta{\~{n}}eda,
  Zald{\'{i}}var, Cuevas, Fausto, and Rodr{\'{i}}guez]{Morales-Castaneda2020}
Morales-Casta{\~{n}}eda, B., Zald{\'{i}}var, D., Cuevas, E., Fausto, F., and
  Rodr{\'{i}}guez, A.
\newblock {A better balance in metaheuristic algorithms: Does it exist?}
\newblock \emph{Swarm and Evolutionary Computation}, 54:\penalty0 100671, may
  2020.
\newblock ISSN 22106502.
\newblock \doi{10.1016/j.swevo.2020.100671}.

\bibitem[Pang et~al.(2020)Pang, Yang, Dong, Xu, Zhu, and Su]{Pang2020}
Pang, T., Yang, X., Dong, Y., Xu, K., Zhu, J., and Su, H.
\newblock {Boosting Adversarial Training with Hypersphere Embedding}.
\newblock \emph{Advances in Neural Information Processing Systems},
  2020-December, feb 2020.
\newblock ISSN 10495258.

\bibitem[Rade \& Moosavi-Dezfooli(2021)Rade and Moosavi-Dezfooli]{Rade2021}
Rade, R. and Moosavi-Dezfooli, S.-M.
\newblock Helper-based adversarial training: Reducing excessive margin to
  achieve a better accuracy vs. robustness trade-off.
\newblock In \emph{ICML 2021 Workshop on Adversarial Machine Learning}, 2021.

\bibitem[Rebuffi et~al.(2021)Rebuffi, Gowal, Calian, Stimberg, Wiles, and
  Mann]{Rebuffi2021}
Rebuffi, S.-A., Gowal, S., Calian, D.~A., Stimberg, F., Wiles, O., and Mann,
  T.~A.
\newblock Fixing data augmentation to improve adversarial robustness.
\newblock \emph{CoRR}, abs/2103.01946, 2021.

\bibitem[Rice et~al.(2020)Rice, Wong, and Kolter]{Rice2020}
Rice, L., Wong, E., and Kolter, Z.
\newblock Overfitting in adversarially robust deep learning.
\newblock In III, H.~D. and Singh, A. (eds.), \emph{Proceedings of the 37th
  International Conference on Machine Learning}, volume 119 of
  \emph{Proceedings of Machine Learning Research}, pp.\  8093--8104. PMLR,
  13--18 Jul 2020.

\bibitem[Said et~al.(2018)Said, Abbasi, Maqbool, Daud, and Aljohani]{Said2018}
Said, A., Abbasi, R.~A., Maqbool, O., Daud, A., and Aljohani, N.~R.
\newblock {CC-GA: A clustering coefficient based genetic algorithm for
  detecting communities in social networks}.
\newblock \emph{Applied Soft Computing}, 63:\penalty0 59--70, feb 2018.
\newblock ISSN 1568-4946.
\newblock \doi{10.1016/J.ASOC.2017.11.014}.

\bibitem[Salman et~al.(2020)Salman, Ilyas, Engstrom, Kapoor, and
  Madry]{DBLP:journals/corr/abs-2007-08489}
Salman, H., Ilyas, A., Engstrom, L., Kapoor, A., and Madry, A.
\newblock Do adversarially robust imagenet models transfer better?
\newblock \emph{CoRR}, abs/2007.08489, 2020.

\bibitem[Sehwag et~al.(2020)Sehwag, Wang, Mittal, and Jana]{Sehwag2020}
Sehwag, V., Wang, S., Mittal, P., and Jana, S.
\newblock {HYDRA: Pruning Adversarially Robust Neural Networks}.
\newblock \emph{Advances in Neural Information Processing Systems},
  2020-December, feb 2020.
\newblock ISSN 10495258.

\bibitem[Sehwag et~al.(2021)Sehwag, Mahloujifar, Handina, Dai, Xiang, Chiang,
  and Mittal]{Sehwag2021}
Sehwag, V., Mahloujifar, S., Handina, T., Dai, S., Xiang, C., Chiang, M., and
  Mittal, P.
\newblock Robust learning meets generative models: Can proxy distributions
  improve adversarial robustness?, 2021.

\bibitem[Sitawarin et~al.(2021)Sitawarin, Chakraborty, and
  Wagner]{Sitawarin2020}
Sitawarin, C., Chakraborty, S., and Wagner, D.
\newblock Sat: Improving adversarial training via curriculum-based loss
  smoothing.
\newblock In \emph{Proceedings of the 14th ACM Workshop on Artificial
  Intelligence and Security}, AISec '21, pp.\  25–36, New York, NY, USA,
  2021. Association for Computing Machinery.
\newblock ISBN 9781450386579.
\newblock \doi{10.1145/3474369.3486878}.

\bibitem[Sridhar et~al.(2021)Sridhar, Sokolsky, Lee, and Weimer]{Sridhar2021}
Sridhar, K., Sokolsky, O., Lee, I., and Weimer, J.
\newblock {Improving Neural Network Robustness via Persistency of Excitation}.
\newblock jun 2021.

\bibitem[Szegedy et~al.(2014)Szegedy, Zaremba, Sutskever, Bruna, Erhan,
  Goodfellow, and Fergus]{szegedy2013intriguing}
Szegedy, C., Zaremba, W., Sutskever, I., Bruna, J., Erhan, D., Goodfellow, I.,
  and Fergus, R.
\newblock Intriguing properties of neural networks.
\newblock In \emph{2nd International Conference on Learning Representations,
  {ICLR} 2014, Banff, AB, Canada, April 14-16, 2014, Conference Track
  Proceedings}, 2014.

\bibitem[Tabak et~al.(2014)Tabak, Takami, Rocha, Cajueiro, and
  Souza]{Tabak2014}
Tabak, B.~M., Takami, M., Rocha, J.~M., Cajueiro, D.~O., and Souza, S.~R.
\newblock {Directed clustering coefficient as a measure of systemic risk in
  complex banking networks}.
\newblock \emph{Physica A: Statistical Mechanics and its Applications},
  394:\penalty0 211--216, jan 2014.
\newblock ISSN 0378-4371.
\newblock \doi{10.1016/J.PHYSA.2013.09.010}.

\bibitem[Wang et~al.(2020)Wang, Zou, Yi, Bailey, Ma, and Gu]{Wang2019}
Wang, Y., Zou, D., Yi, J., Bailey, J., Ma, X., and Gu, Q.
\newblock Improving adversarial robustness requires revisiting misclassified
  examples.
\newblock In \emph{International Conference on Learning Representations}, 2020.

\bibitem[Wong et~al.(2020)Wong, Rice, and Kolter]{Wong2020}
Wong, E., Rice, L., and Kolter, J.~Z.
\newblock Fast is better than free: Revisiting adversarial training.
\newblock In \emph{8th International Conference on Learning Representations,
  {ICLR} 2020, Addis Ababa, Ethiopia, April 26-30, 2020}. OpenReview.net, 2020.

\bibitem[Wu et~al.(2020)Wu, Xia, and Wang]{Wu2020a}
Wu, D., Xia, S.~T., and Wang, Y.
\newblock {Adversarial Weight Perturbation Helps Robust Generalization}.
\newblock \emph{Advances in Neural Information Processing Systems},
  2020-December, apr 2020.
\newblock ISSN 10495258.

\bibitem[Xie et~al.(2019)Xie, Zhang, Zhou, Bai, Wang, Ren, and
  Yuille]{Xie_2019_CVPR}
Xie, C., Zhang, Z., Zhou, Y., Bai, S., Wang, J., Ren, Z., and Yuille, A.~L.
\newblock Improving transferability of adversarial examples with input
  diversity.
\newblock In \emph{Proceedings of the IEEE/CVF Conference on Computer Vision
  and Pattern Recognition (CVPR)}, June 2019.

\bibitem[Yao et~al.(2019)Yao, Gholami, Xu, Keutzer, and Mahoney]{Yao_2019_CVPR}
Yao, Z., Gholami, A., Xu, P., Keutzer, K., and Mahoney, M.~W.
\newblock Trust region based adversarial attack on neural networks.
\newblock In \emph{Proceedings of the IEEE/CVF Conference on Computer Vision
  and Pattern Recognition (CVPR)}, June 2019.

\bibitem[Zhang et~al.(2019{\natexlab{a}})Zhang, Zhang, Lu, Zhu, and
  Dong]{Zhang2019a}
Zhang, D., Zhang, T., Lu, Y., Zhu, Z., and Dong, B.
\newblock {You Only Propagate Once: Accelerating Adversarial Training via
  Maximal Principle}.
\newblock \emph{Advances in Neural Information Processing Systems}, 32, may
  2019{\natexlab{a}}.
\newblock ISSN 10495258.

\bibitem[Zhang et~al.(2019{\natexlab{b}})Zhang, Yu, Jiao, Xing, Ghaoui, and
  Jordan]{Zhang2019}
Zhang, H., Yu, Y., Jiao, J., Xing, E.~P., Ghaoui, L.~E., and Jordan, M.~I.
\newblock {Theoretically Principled Trade-off between Robustness and Accuracy}.
\newblock \emph{36th International Conference on Machine Learning, ICML 2019},
  2019-June:\penalty0 12907--12929, jan 2019{\natexlab{b}}.

\bibitem[Zhang et~al.(2020)Zhang, Xu, Han, Niu, Cui, Sugiyama, and
  Kankanhalli]{Zhang2020a}
Zhang, J., Xu, X., Han, B., Niu, G., Cui, L., Sugiyama, M., and Kankanhalli, M.
\newblock Attacks which do not kill training make adversarial learning
  stronger.
\newblock In III, H.~D. and Singh, A. (eds.), \emph{Proceedings of the 37th
  International Conference on Machine Learning}, volume 119 of
  \emph{Proceedings of Machine Learning Research}, pp.\  11278--11287. PMLR,
  13--18 Jul 2020.

\bibitem[Zhang et~al.(2021)Zhang, Zhu, Niu, Han, Sugiyama, and
  Kankanhalli]{Zhang2020}
Zhang, J., Zhu, J., Niu, G., Han, B., Sugiyama, M., and Kankanhalli, M.
\newblock Geometry-aware instance-reweighted adversarial training.
\newblock In \emph{International Conference on Learning Representations}, 2021.

\bibitem[Zheng et~al.(2019)Zheng, Chen, and Ren]{Zheng_Chen_Ren_2019}
Zheng, T., Chen, C., and Ren, K.
\newblock Distributionally adversarial attack.
\newblock \emph{Proceedings of the AAAI Conference on Artificial Intelligence},
  33\penalty0 (01):\penalty0 2253--2260, Jul. 2019.
\newblock \doi{10.1609/aaai.v33i01.33012253}.

\end{thebibliography}
\bibliographystyle{icml2022}

\clearpage
\renewcommand{\thesection}{\Alph{section}}
\renewcommand{\thesubsection}{\arabic{subsection}}
\setcounter{section}{0}
\crefalias{section}{appendix}
\begin{subappendices}

\section*{Appendix}
In this appendix, we provide the following additional information regarding the experimental results and the behavior of the proposed approach during the search for ACG.
\begin{itemize}
    \item An example of desirable search behavior. (\Cref{appendix:exampleDiversificationAndIntensification})
    \item Transitions of the best objective values and the effect of increasing the iterations.
    (\Cref{ComparisonTheBestLoss})
    \item The effect of random restarts on the performance of ACG.
    (\Cref{comparison_500iter})
    \item A discussion of the diversification and intensification of APGD search.
    (\Cref{appendix:section:Diversification and Intensification of APGD search})
    \item The effect of the objective function on the performance of ACG.
    (\Cref{results_of_dlr_loss})
    \item The effect of the operation conducted to render $\beta$ non-negative, which is known to perform better for a general nonlinear optimization.
    (\Cref{appendix:positivebeta_conjugate})
    \item Specifications of the computational environments used in our experiments.
    (\Cref{appendix:experiment_environments})
    \item Differences in the ASR among different formulas used to calculate $\beta$.
    (\Cref{appendix:OtherConjugateFormulation})
    \item Differences in the search behavior of APGD and ACG for the model in which the ASR of ACG was lower than that of APGD and the other models.
    (\Cref{appendix:addepalli_analysis})
    \item A definition of \measurement on arbitrary bounded distance spaces.
    (\Cref{generalized_DI})
    \item The relationship between distributions of the point clouds and \measurementNoSpace.
    (\Cref{additional_examples_of_DI})
\end{itemize}

\section{Related works}
\label{appendix:related_works}
\begin{figure}[tb]
  \centering
      \includegraphics[valign=m,width=\linewidth]{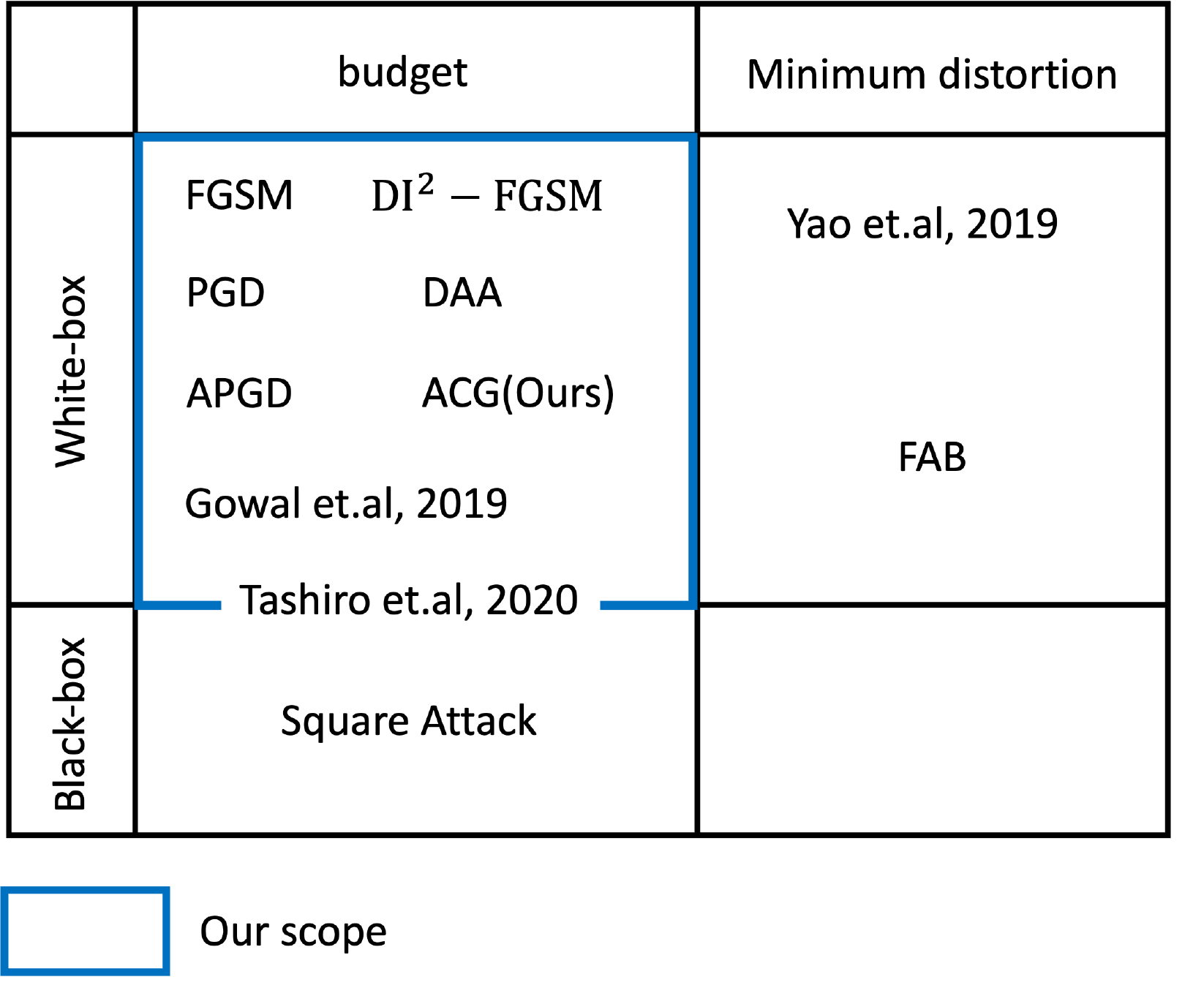}
   \caption{
   Classification of previous works.
   }
   \label{related_works}
\end{figure}

We classify some previous works to clarify the position of the proposed method.
\Cref{related_works} shows an overview.

"Budget" is a type of adversarial attacks using the formulation described in \Cref{preliminaries}. \cite{DBLP:journals/corr/abs-1910-09338,Xie_2019_CVPR,Zheng_Chen_Ren_2019} are examples of this type of attacks.

"Minimum distortion" is another type of adversarial attack which generates minimally distorted adversarial examples by minimizing the norm of adversarial perturbation.
Formally, this problem is described as follows.
$$
\minop_{x\in\D} d(x_{\rm orig}, x)\textrm{ ~s.t.~} \argmax_{k=1,\dots K}~g_k(x)\neq c
$$.
\cite{Yao_2019_CVPR,croce2020minimally} are examples of this type of attacks.

In the case of white-box attacks, we can assume that all information about the model to be attacked is known the attacker, including its weights and gradients.

In the case of black-box attacks, only use the output of target models can be used.
The proposed method, ACG, is a white-box attack that uses a budget formulation.

\section{Examples of Diversification and Intensification}
\label{appendix:exampleDiversificationAndIntensification}
\begin{figure*}[tb]
    \vskip 0.2in
    \centering
    \includegraphics[width=\linewidth]{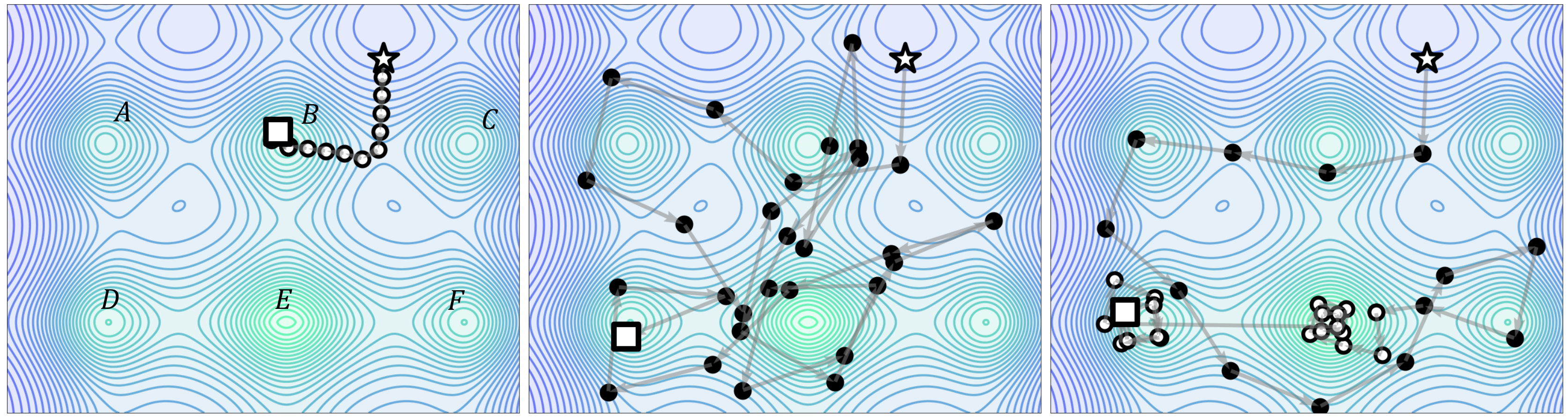}
    \caption{
    Examples of searches for multimodal functions. The initial search point is shown by the white star, and the search ended at a white square. The circles represent the search points. The black circle indicates that the search was diversified, and the white circle indicates an intense search. "A" to "F" on the left figure refers to the local optima, and "E" shows the global optima. The examples on the left and center are those of failed searches that were not able to find the global optima owing to a lack of diversification or intensification, respectively. The example on the right shows a search that has successfully found the global optima owing to an appropriate balance of diversification and intensification, which is our aim in the present work.
    }
    \label{fig:example_of_diversification}
    \vskip -0.2in
\end{figure*}

In this section, we detail the balance between diversification and intensification that we aim to realize.

Most white-box adversarial attacks are formulated as optimization problems where the objective function is nonlinear, nonconvex, and multimodal.
We observe the diversification and intensification of the search to find the optimal solution by using the following multimodal function.

\begin{align}
\notag f(x, y) = & -10 \exp \left(-0.2 \sqrt{\frac{1}{2} (x^4 + y^2)} \right)\\
& +\exp \left( \frac{1}{2} \left(\cos (2 \pi x) + \cos (2 \pi y) \right) \right)  .
\label{function:multimodal}
\end{align}

\Cref{fig:example_of_diversification} shows examples of an intensified search, a diversified search, and an appropriate search exhibiting the proper balance of diversification and intensification.
Six local solutions may be observed from the contour lines of the function in the figure, of which the local solution at position "E" is the global optimum.

In the figure on the left, the search (white circles) was intensified in the local solution near the initial point.
In the middle figure, the search (black circles) was diversified even when reaching the neighborhood of the local solution.
Therefore, it appears to be impossible to reach an optimal solution if the search is excessively intensified or diversified.
In the figure on the right, it may be observed that the search changed from a focus on diversification to a focus on intensification and that the search reached the neighbor of the global optima through a diversified search (black circles) and converged to the optimal solution through an intensive search (white circles).
Thus, for multimodal functions, a balance between diversification and intensification of the search is necessary when searching an entire feasible region to finding an optimum.
The right side of Figure 6 shows our intended search, in which the balance between diversification and intensification is properly controlled.

\section{\textbf{ACG vs. APGD:} Comparison of the best objective values}
\label{ComparisonTheBestLoss}
\begin{figure}[tb]
    \vskip 0.2in
    \centering
    \includegraphics[width=\linewidth]{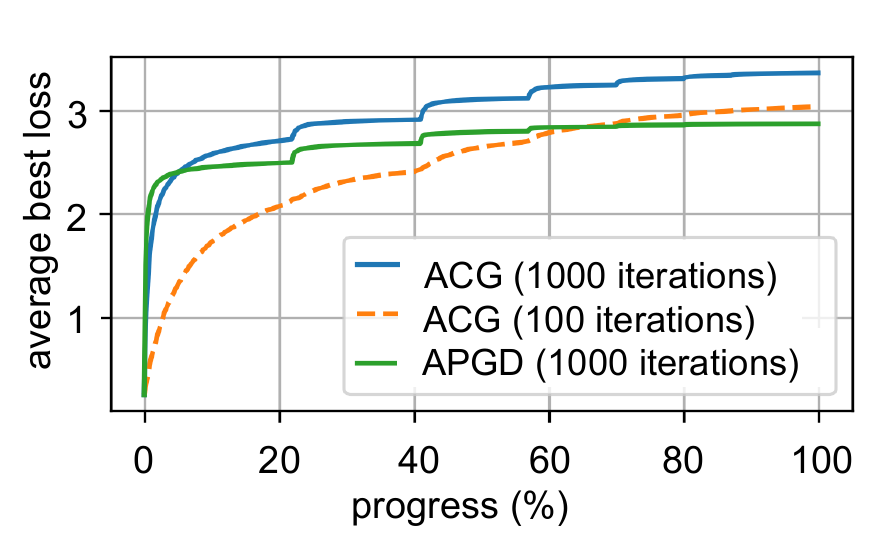}
    \caption{Transitions of the best objective values averaged over 10,000 images. Here, "progress" means the percentage of the iterations to the total iterations.
    }
    \label{appendix:bestloss}
    \vskip -0.2in
\end{figure}
From \Cref{appendix:bestloss}, we predicted that ACG (100 iterations) could still improve the objective value because it improves the best objective value to a relatively large degree even at the end of the search (dashed line, orange). We then investigated the search performance of ACG in terms of the best objective values.
\Cref{appendix:bestloss} shows the transitions of the best loss of ACG after 100 iterations, and those of APGD and ACG after 1000 iterations. The x-axis represents the percentage of the iterations to $N_\mathrm{iter}$. The best loss at iteration $k$ is defined as $\displaystyle\iter{{\rm best~loss}}{k}=\underset{i\leq k}{\max} \ L\left(g(\iter{\bs{x}}{i}), c\right)$.
As expected, ACG (1000 iterations) exhibited higher objective values than ACG (100 iterations), whereas ACG (1000 iterations) improved on the best objective values even at the end of the search. By contrast, the improvement of the best objective values of APGD (1000 iterations) was small at the end of the attack compared to ACG (100 iterations, 1000 iterations).
To summarize these results, ACG may be considered a more promising algorithm than APGD because, in contrast to APGD,  it can significantly improve on its best loss even at the end of the search.

\section{Effects of random restarts for the performance of ACG}
\label{comparison_500iter}
In this section, we compare the results of APGD with 5 restarts for 100 iterations, ACG with 5 restarts for 100 iterations, and ACG with 1 restart for 500 iterations.
\Cref{table:comparison_500iter_cifar10,table:comparison_500iter_cifar100,table:comparison_500iter_imagenet} show that ACG(5), which is randomized by the restarts, and ACG-500iter, which is deterministic, achieved almost the same ASR. This result suggests that the search diversity of ACG does not depend on the random sampling of the initial points, but that the update direction itself has the property of a diverse search compared to APGD.

\section{Diversification and Intensification of APGD Search}
\label{appendix:section:Diversification and Intensification of APGD search}
Herein, we discuss the diversification and intensification of APGD.
APGD is a PGD-based adversarial attack that achieves a higher ASR than previous SOTA methods by gradually reducing the step size, as shown in \Cref{sec:Step Size Selection}.
APGD introduced this step size reduction to gradually switch from exploration to exploitation.
Note that ``exploration'' and ``exploitation'' are synonymous with ``diversification'' and ``intensification'' as we used the terms in this work.
However, to the best of out knowledge, whether APGD switches from diversification to intensification has not been verified.

\Cref{clustering_coef_APGD_and_ACG} in \Cref{subsection:ComparisonOfApgdAndAcgAboutDI} show that the DIs of APGD gradually decreased during the search for 100 iterations.
These results indicate that APGD can switch from diversification to intensification in terms of DI, and thus the motivation for the step size selection was achieved.
However, the DI value of APGD was smaller than that of ACG.
We discuss the reason for this based on the results described in \Cref{subsection:effectOfConjugate}.
From the image at the top of \Cref{fig:move_distance}, it may be observed that the projected distance of the PGD-based updated search point to the feasible region was larger than that of the CG-based search point.
This means that with APGD, the search points are close to the boundary of the region and are updated toward the outside of the boundary.
Because the PGD-based method tends to update toward the same local optimum, doing so is natural when there is a local optimum outside the feasible region.
As a result, a small distance was induced between two successive search points (as shown at the bottom of \Cref{fig:move_distance}), dense search points, and small DI values.

Because ACG, which exhibited a more variable DI, delivers a better performance than APGD, we expect that higher-performing adversarial attacks will be developed in future research via sophisticated control of the DI transition.

\section{DLR loss}
\label{results_of_dlr_loss}
In this section, we compare the ASR under the experimental setup of \Cref{sec:experiments}; however, we used the DLR loss proposed in \cite{croce2020reliable} as the objective function instead of the CW loss in  \Cref{sec:experiments}.
\Cref{table:dlr_center_random_cifar10andimagenet,table:dlr_center_random_cifar100} show that the ASR of ACG was higher than that of APGD for all 64 models when the objective function was the DLR loss.

\section{Analysis of effect of the assumption of $\beta_{{\rm HS}} \geq 0$}
\label{appendix:positivebeta_conjugate}
\begin{figure*}[t]
  \centering
  \begin{tabular}{wc{0.1in} ccc} 
  &&& \\
    & \cite{Ding2019} & \cite{Rebuffi2021}  & \cite{Carmon2019} \\ 
\begin{tabular}{c}
 \rotatebox[origin=c]{90}{Diversity Index}
  \end{tabular}
    &
      \centering
      \scalebox{1.0}{\includegraphics[valign=m,width=0.3\linewidth]{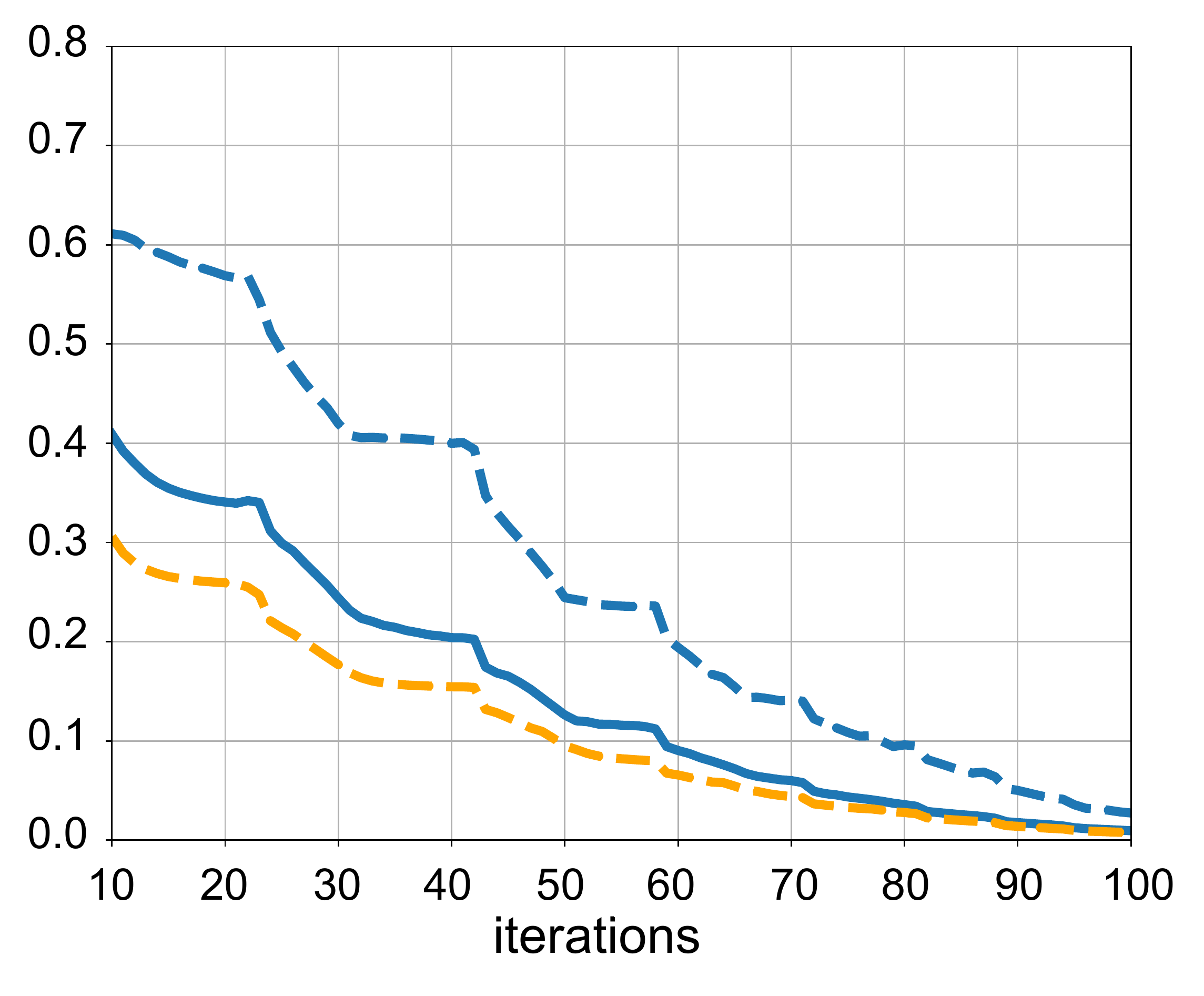}}
    &
      \centering
      \scalebox{1.0}{\includegraphics[valign=m,width=0.3\linewidth]{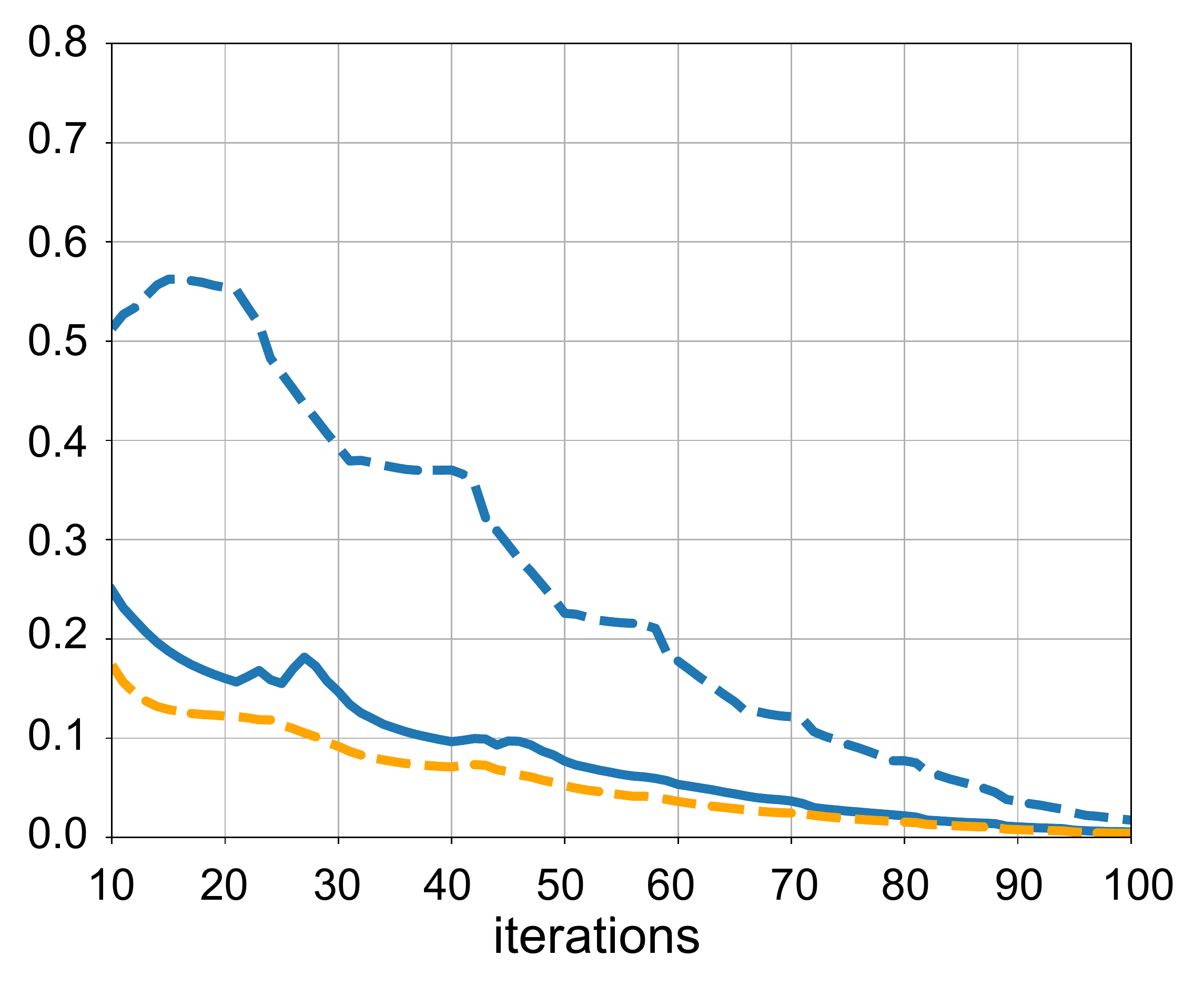}}
    &
      \centering
      \scalebox{1.0}{\includegraphics[valign=m,width=0.3\linewidth]{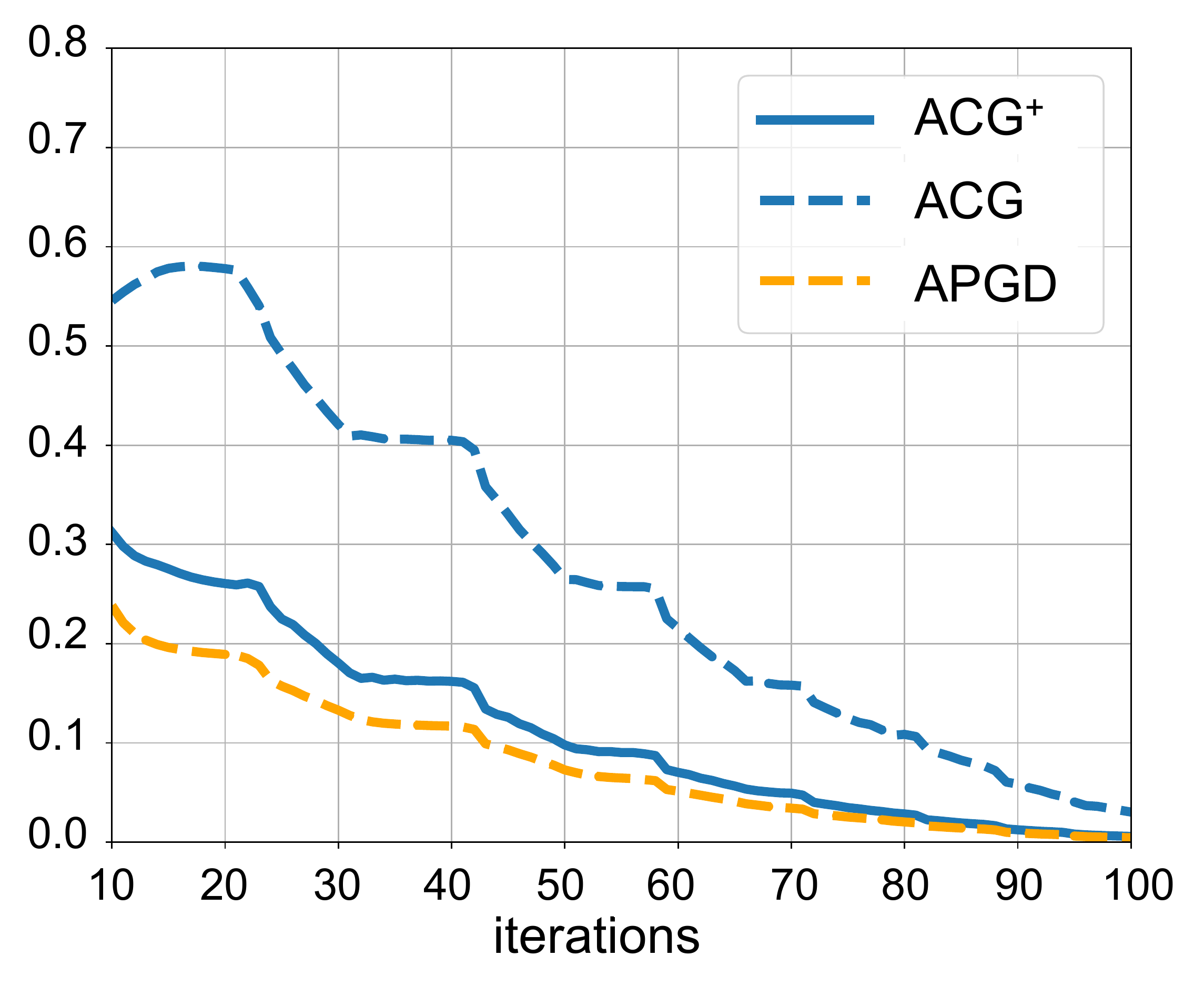}}
  \end{tabular}
   \caption{Comparison of \measurement of $\textrm{ACG}^+$, ACG, and APGD. The CIFAR-10 dataset was used for this comparison.
   }
   \label{figure:positive_beta} 
\end{figure*}

When discussing the convergence of the conjugate gradient method, it has occasionally been assumed that $\beta_{\textrm HS}\geq 0$ \cite{Hager2006}. In addition, some studies have suggested that it is better to assume $\beta_{\textrm HS} \geq 0$ is preferable in practice.
In this section, we investigate the effect of assuming $\beta_{\textrm HS} \geq 0$ by comparing the behavior of ACG without any assumptions on $\beta_{\textrm HS}$.
To render $\beta_{\textrm HS}$ be non-negative, we can obtain the following.
\begin{align}\label{eq:positive_beta}
\beta_{\textrm HS}^+ := \max (\beta_{\textrm HS}, 0).
\end{align}
We call ACG using $\beta_{\textrm HS}^+$ as $\textrm{ACG}^+$, as determined using ~\Cref{eq:positive_beta}.
\Cref{table:appendix_positivebeta_conjugate} show that the ASR of $\textrm{ACG}^+$ was equal to or less than that of APGD, suggesting a decrease in the search performance of $\textrm{ACG}^+$.
In addition, the percentage of iterations where $\beta_{\textrm{HS}}^+=0$ was 30\% to 40\% of all 495 iterations, excluding iteration 0, where the steepest descent occured(see~\Cref{appendix:table:theRatioOfPositiveBeta}).
That is, $\textrm{ACG}^+$ updated in the same way as APGD without a momentum update method once every three iterations.
In addition, from \Cref{figure:positive_beta}, it may be observed that $\textrm{ACG}^+$ diversified the search less than ACG.
These results show that operations that make $\beta_{\textrm{HS}}$ nonnegative, such as \Cref{eq:positive_beta}, are unsuitable for this problem. Applying the operation to ACG limits the diversification performance, which is one of the strengths of ACG.

\begin{table}[t]
\caption{The average percentage of the iterations in which $\beta_{\textrm HS}^+$=0 among all 495 iterations. The CIFAR-10 dataset was used for this experiment, and the diameter of the feasible region $\varepsilon$ was $8/255$.
}
\label{appendix:table:theRatioOfPositiveBeta}
\vskip 0.15in
\centering
\begin{tabular}{c|c}
\toprule
\textbf{paper} & \textbf{Ratio} of $\beta_{\textrm HS}^+$ is $0$  \\ \midrule
\cite{Ding2019}    & 40.59\%   \\
\cite{Carmon2019}  & 30.41\%  \\
\cite{Rebuffi2021} & 38.57\% \\ \bottomrule
\end{tabular}
\end{table}
\vskip -0.1in
\begin{table*}
\centering
\caption{ASR of $\textrm{ACG}^+$ and APGD:
    The dataset is CIFAR-10, and $\varepsilon=\frac{8}{255}$.
}
\vskip 0.15in
\begin{tabular}{ c | c | cccc || c}
\toprule
CIFAR-10 ($\varepsilon=8/255$) & & \multicolumn{4}{c}{Attack Success Rate}\\ \midrule
\bf{paper} & \bf{Architecture} & \bf{APGD(1)} & \bf{$\textrm{ACG}^+$(1)} & \bf{APGD(5)} & \bf{$\textrm{ACG}^+$(5)} & \bf{diff} \\ 
\hline\hline
\cite{Rebuffi2021} & PreActResNet-18 & 42.73 & 42.71 & \bf{42.91} & \underline{42.88} & -0.03\\
\cite{Carmon2019} & WideResNet-28-10 & 39.38 & 39.28 & \bf{39.59} & \underline{39.51} & -0.08\\
\cite{Ding2019} & WideResNet-28-4 & 48.73 & 49.66 & \underline{49.67} & \bf{50.68} & 1.01\\
\bottomrule
\end{tabular}
\vskip -0.1in
\label{table:appendix_positivebeta_conjugate}
\end{table*}
\begin{table*}[tb]
\centering
\caption{The ASR of APGD with 5 restarts, ACG with 5 restarts, and ACG with 1 restart for 500 iterations. The CIFAR-10 dataset was used with $\varepsilon=\frac{8}{255}$. The highest ASR is in bold, and the second is underlined. APGD($N$) refers to APGD with $N$-times the initial point selection. The meanings of the other columns are the same. 
\emph{diff} is the difference between APGD(5) and $\max\{$ACG(5), ACG-500iter$\}$.}

\label{table:comparison_500iter_cifar10}
\vskip 0.15in
\begin{tabular}{ c | c | ccc || c}\toprule
CIFAR-10 ($\varepsilon=8/255$) & & \multicolumn{3}{c}{Attack Success Rate}\\ \midrule
\bf{paper} & \bf{Architecture} & \bf{APGD(5)} & \bf{ACG(5)} & \bf{ACG-500iter} & \bf{diff} \\ 
 \hline\hline
\cite{Rade2021} & PreActResNet-18 & 42.46 & \underline{42.65} & \bf{42.67} & 0.21\\
\cite{Rade2021} & PreActResNet-18 & 41.65 & \bf{42.12} & \underline{42.07} & 0.47\\
\cite{Rebuffi2021} & PreActResNet-18 & 42.91 & \underline{43.15} & \bf{43.17} & 0.26\\
\cite{andriushchenko2020square} & PreActResNet-18 & 53.82 & \bf{54.90} & \underline{54.81} & 1.08\\
\cite{Sehwag2021} & ResNet-18 & 43.91 & \bf{44.79} & \underline{44.53} & 0.88\\
\cite{Chen2020a} & ResNet-50 & 48.08 & \underline{48.28} & \bf{48.29} & 0.21\\
\cite{Wong2020} & ResNet-50 & 54.26 & \bf{55.44} & \underline{55.34} & 1.18\\
\cite{engstrom2019} & ResNet-50 & 48.08 & \bf{49.25} & \underline{49.08} & 1.17\\
\cite{Rebuffi2021} & WideResNet-106-16 & 34.71 & \bf{35.03} & \underline{34.98} & 0.32\\
\cite{Carmon2019} & WideResNet-28-10 & 39.59 & \bf{40.03} & \underline{39.98} & 0.44\\
\cite{Gowal2020} & WideResNet-28-10 & 36.45 & \underline{36.90} & \bf{36.96} & 0.51\\
\cite{Hendrycks2019} & WideResNet-28-10 & 43.82 & \underline{44.36} & \bf{44.37} & 0.55\\
\cite{Rade2021} & WideResNet-28-10 & 38.64 & \bf{38.87} & \underline{38.79} & 0.23\\
\cite{Rebuffi2021} & WideResNet-28-10 & 38.47 & \bf{38.80} & \underline{38.77} & 0.33\\
\cite{Sehwag2020} & WideResNet-28-10 & 41.93 & \underline{42.41} & \bf{42.48} & 0.55\\
\cite{Sridhar2021} & WideResNet-28-10 & 39.45 & \underline{39.85} & \bf{39.88} & 0.43\\
\cite{Wang2019} & WideResNet-28-10 & 42.15 & \bf{42.57} & \underline{42.53} & 0.42\\
\cite{Wu2020a} & WideResNet-28-10 & 39.56 & \underline{39.70} & \bf{39.72} & 0.16\\
\cite{Zhang2020} & WideResNet-28-10 & 39.98 & \bf{40.25} & \underline{40.24} & 0.27\\
\cite{Ding2019} & WideResNet-28-4 & 49.67 & \bf{55.77} & \underline{55.32} & 6.10\\
\cite{Cui2020} & WideResNet-34-10 & 46.41 & \underline{46.90} & \bf{46.90} & 0.49\\
\cite{Huang2020} & WideResNet-34-10 & 46.19 & \bf{46.72} & \underline{46.67} & 0.53\\
\cite{Rade2021} & WideResNet-34-10 & 36.46 & \bf{36.83} & \underline{36.77} & 0.37\\
\cite{Sehwag2021} & WideResNet-34-10 & 39.58 & \underline{40.18} & \bf{40.18} & 0.60\\
\cite{Sitawarin2020} & WideResNet-34-10 & 47.23 & \underline{48.02} & \bf{48.05} & 0.82\\
\cite{Wu2020a} & WideResNet-34-10 & 43.36 & \underline{43.60} & \bf{43.62} & 0.26\\
\cite{Zhang2019a} & WideResNet-34-10 & 53.08 & \underline{54.15} & \bf{54.17} & 1.09\\
\cite{Zhang2019} & WideResNet-34-10 & 46.65 & \underline{47.18} & \bf{47.19} & 0.54\\
\cite{Zhang2020a} & WideResNet-34-10 & 45.68 & \underline{46.12} & \bf{46.13} & 0.45\\
\cite{chen2020Efficient} & WideResNet-34-10 & 47.58 & \bf{48.00} & \underline{47.99} & 0.42\\
\cite{Sridhar2021} & WideResNet-34-15 & 38.90 & \underline{39.15} & \bf{39.22} & 0.32\\
\cite{Cui2020} & WideResNet-34-20 & 45.88 & \bf{46.23} & \underline{46.14} & 0.35\\
\cite{Gowal2020} & WideResNet-34-20 & 42.65 & \underline{42.86} & \bf{42.91} & 0.26\\
\cite{Pang2020} & WideResNet-34-20 & 44.75 & \underline{45.33} & \bf{45.34} & 0.59\\
\cite{Rice2020} & WideResNet-34-20 & 44.92 & \underline{45.69} & \bf{45.73} & 0.81\\
\cite{Huang2021} & WideResNet-34-R & 37.33 & \underline{37.79} & \bf{37.90} & 0.57\\
\cite{Huang2021} & WideResNet-34-R & 36.27 & \underline{36.76} & \bf{36.81} & 0.54\\
\cite{Gowal2020} & WideResNet-70-16 & 33.42 & \underline{33.70} & \bf{33.82} & 0.40\\
\cite{Gowal2020} & WideResNet-70-16 & 42.12 & \bf{42.45} & \underline{42.40} & 0.33\\
\cite{Gowal2021} & WideResNet-70-16 & 32.57 & \bf{33.04} & \underline{32.95} & 0.47\\
\cite{Rebuffi2021} & WideResNet-70-16 & 35.04 & \bf{35.27} & \underline{35.19} & 0.23\\
\cite{Rebuffi2021} & WideResNet-70-16 & 32.46 & \bf{32.75} & \underline{32.69} & 0.29\\

\bottomrule
\end{tabular}
\vskip -0.1in
\end{table*}

\begin{table*}
\centering
\caption{The ASR of APGD with 5 restarts, ACG with 5 restarts, and ACG with 1 restart for 500 iterations. The CIFAR-100 dataset with $\varepsilon=\frac{8}{255}$ was used. The highest ASR is in bold, and the second is underlined. APGD($N$) refers to APGD with $N$-times the initial point selection. The meanings of the other columns are the same. 
\emph{diff} is the difference between APGD(5) and $\max\{$ACG(5), ACG-500iter$\}$.
}
\label{table:comparison_500iter_cifar100}
\vskip 0.15in
\begin{tabular}{ c | c | ccc || c}\toprule
CIFAR-100 ($\varepsilon=8/255$) & & \multicolumn{3}{c}{Attack Success Rate}\\ \midrule

\bf{paper} & \bf{Architecture} & \bf{APGD(5)} & \bf{ACG(5)} & \bf{ACG-500iter} & \bf{diff} \\ \hline\hline
\cite{Addepalli2021} & PreActResNet-18 & 72.25 & \bf{72.47} & \underline{72.36} & 0.22\\
\cite{Rade2021} & PreActResNet-18 & 70.55 & \bf{70.86} & \underline{70.77} & 0.31\\
\cite{Rebuffi2021} & PreActResNet-18 & 70.93 & \bf{71.29} & \underline{71.21} & 0.36\\
\cite{Rice2020} & PreActResNet-18 & 79.99 & \bf{80.63} & \underline{80.55} & 0.64\\
\cite{Hendrycks2019} & WideResNet-28-10 & 69.50 & \bf{70.51} & \underline{70.44} & 1.01\\
\cite{Rebuffi2021} & WideResNet-28-10 & 66.67 & \bf{67.27} & \underline{67.16} & 0.60\\
\cite{Addepalli2021} & WideResNet-34-10 & \bf{68.74} & 68.52 & \underline{68.59} & -0.15\\
\cite{Chen2021} & WideResNet-34-10 & 68.36 & \bf{68.77} & \underline{68.68} & 0.41\\
\cite{Cui2020} & WideResNet-34-10 & 69.87 & \bf{70.33} & \underline{70.32} & 0.46\\
\cite{Cui2020} & WideResNet-34-10 & 72.15 & \underline{72.56} & \bf{72.64} & 0.49\\
\cite{Sitawarin2020} & WideResNet-34-10 & 73.43 & \bf{74.27} & \underline{74.09} & 0.84\\
\cite{Wu2020a} & WideResNet-34-10 & 69.32 & \bf{70.11} & \underline{69.97} & 0.79\\
\cite{chen2020Efficient} & WideResNet-34-10 & 71.96 & \underline{72.18} & \bf{72.19} & 0.23\\
\cite{Cui2020} & WideResNet-34-20 & 68.72 & \bf{69.13} & \underline{69.08} & 0.41\\
\cite{Gowal2020} & WideResNet-70-16 & 61.55 & \bf{62.19} & \underline{62.15} & 0.64\\
\cite{Gowal2020} & WideResNet-70-16 & 69.04 & \bf{69.43} & \underline{69.35} & 0.39\\
\cite{Rebuffi2021} & WideResNet-70-16 & 64.17 & \bf{64.77} & \underline{64.61} & 0.60\\
\bottomrule
\end{tabular}
\vskip -0.1in
\end{table*}

\begin{table*}
\centering
\caption{The ASR of APGD with 5 restarts, ACG with 5 restarts, and ACG with 1 restart for 500 iterations. The ImageNet dataset with $\varepsilon=\frac{4}{255}$ is used. The highest ASR is in bold, and the second is underlined. APGD($N$) refers to APGD with $N$-times the initial point selection. The meanings of the other columns are the same. 
\emph{diff} is the difference between APGD(5) and $\max\{$ACG(5), ACG-500iter$\}$.
}
\label{table:comparison_500iter_imagenet}
\vskip 0.15in
\begin{tabular}{ c | c | ccc || c}\toprule
ImageNet ($\varepsilon=4/255$) & & \multicolumn{3}{c}{Attack Success Rate}\\ \midrule

\bf{paper} & \bf{Architecture} & \bf{APGD(5)} & \bf{ACG(5)} & \bf{ACG-500iter} & \bf{diff} \\ \hline\hline
\cite{DBLP:journals/corr/abs-2007-08489} & ResNet-18 & 73.00 & \bf{73.72} & \underline{73.56} & 0.72\\
\cite{DBLP:journals/corr/abs-2007-08489} & ResNet-50 & 62.86 & \bf{63.70} & \underline{63.54} & 0.84\\
\cite{Wong2020} & ResNet-50 & 71.70 & \bf{71.94} & \underline{71.92} & 0.24\\
\cite{engstrom2019} & ResNet-50 & 67.86 & \bf{68.60} & \underline{68.58} & 0.74\\
\cite{DBLP:journals/corr/abs-2007-08489} & WideResNet-50-2 & 58.96 & \bf{59.92} & \underline{59.82} & 0.96\\

\bottomrule
\end{tabular}
\vskip -0.1in
\end{table*}
\begin{table*}
\centering
\vskip -1.5in
\caption{
Comparison of the ASR of APGD and ACG with DLR loss as the objective function.
Dataset: CIFAR10($\varepsilon=\frac{8}{255}$), ImageNet($\varepsilon=\frac{4}{255}$)
}
\label{table:dlr_center_random_cifar10andimagenet}
\vskip 0.15in
\scalebox{0.9}{
\begin{tabular}{ c | c | c | cc || c}
\toprule
CIFAR-10 ($\varepsilon=8/255$) & & & \multicolumn{2}{c}{Attack Success Rate}\\ \hline
\bf{paper} & \bf{Architecture} & \bf{clean acc} &\bf{APGD(5)} & \bf{ACG(5)} & \bf{diff} \\ \hline\hline
\cite{Rade2021} & PreActResNet-18 & 89.02 & \underline{41.61} & \bf{42.13} & 0.52\\
\cite{Rade2021} & PreActResNet-18 & 86.86 & \underline{42.45} & \bf{42.72} & 0.27\\
\cite{Rebuffi2021} & PreActResNet-18 & 83.53 & \underline{42.88} & \bf{43.22} & 0.34\\
\cite{Rice2020} & PreActResNet-18 & 85.34 & \underline{44.25} & \bf{46.06} & 1.81\\
\cite{andriushchenko2020square} & PreActResNet-18 & 79.84 & \underline{52.99} & \bf{55.38} & 2.39\\
\cite{Sehwag2021} & ResNet-18 & 84.59 & \underline{43.38} & \bf{45.11} & 1.73\\
\cite{Chen2020a} & ResNet-50 & 86.04 & \underline{47.82} & \bf{48.35} & 0.53\\
\cite{Wong2020} & ResNet-50 & 83.34 & \underline{53.25} & \bf{55.72} & 2.47\\
\cite{engstrom2019} & ResNet-50 & 87.03 & \underline{47.36} & \bf{49.82} & 2.46\\
\cite{Rebuffi2021} & WideResNet-106-16 & 88.50 & \underline{34.67} & \bf{35.19} & 0.52\\
\cite{Carmon2019} & WideResNet-28-10 & 89.69 & \underline{39.35} & \bf{40.14} & 0.79\\
\cite{Gowal2020} & WideResNet-28-10 & 89.48 & \underline{36.29} & \bf{37.00} & 0.71\\
\cite{Hendrycks2019} & WideResNet-28-10 & 87.11 & \underline{43.03} & \bf{44.75} & 1.72\\
\cite{Rade2021} & WideResNet-28-10 & 88.16 & \underline{38.62} & \bf{39.00} & 0.38\\
\cite{Rebuffi2021} & WideResNet-28-10 & 87.33 & \underline{38.37} & \bf{39.12} & 0.75\\
\cite{Sehwag2020} & WideResNet-28-10 & 88.98 & \underline{41.80} & \bf{42.56} & 0.76\\
\cite{Sridhar2021} & WideResNet-28-10 & 89.46 & \underline{39.14} & \bf{40.06} & 0.92\\
\cite{Wang2019} & WideResNet-28-10 & 87.50 & \underline{41.52} & \bf{43.08} & 1.56\\
\cite{Wu2020a} & WideResNet-28-10 & 88.25 & \underline{39.50} & \bf{39.86} & 0.36\\
\cite{Zhang2020} & WideResNet-28-10 & 89.36 & \underline{39.79} & \bf{40.57} & 0.78\\
\cite{Ding2019} & WideResNet-28-4 & 84.36 & \underline{49.70} & \bf{56.00} & 6.30\\
\cite{Cui2020} & WideResNet-34-10 & 88.22 & \underline{44.28} & \bf{46.92} & 2.64\\
\cite{Huang2020} & WideResNet-34-10 & 83.48 & \underline{45.76} & \bf{46.93} & 1.17\\
\cite{Rade2021} & WideResNet-34-10 & 91.47 & \underline{36.40} & \bf{36.93} & 0.53\\
\cite{Sehwag2021} & WideResNet-34-10 & 86.68 & \underline{39.22} & \bf{40.61} & 1.39\\
\cite{Sitawarin2020} & WideResNet-34-10 & 86.84 & \underline{46.79} & \bf{48.78} & 1.99\\
\cite{Wu2020a} & WideResNet-34-10 & 85.36 & \underline{43.36} & \bf{43.72} & 0.36\\
\cite{Zhang2019a} & WideResNet-34-10 & 87.20 & \underline{52.59} & \bf{54.65} & 2.06\\
\cite{Zhang2019} & WideResNet-34-10 & 84.92 & \underline{46.51} & \bf{47.27} & 0.76\\
\cite{Zhang2020a} & WideResNet-34-10 & 84.52 & \underline{45.44} & \bf{46.26} & 0.82\\
\cite{chen2020Efficient} & WideResNet-34-10 & 85.32 & \underline{47.32} & \bf{48.24} & 0.92\\
\cite{Sridhar2021} & WideResNet-34-15 & 86.53 & \underline{38.65} & \bf{39.27} & 0.62\\
\cite{Cui2020} & WideResNet-34-20 & 88.70 & \underline{44.68} & \bf{46.39} & 1.71\\
\cite{Gowal2020} & WideResNet-34-20 & 85.64 & \underline{42.57} & \bf{43.02} & 0.45\\
\cite{Pang2020} & WideResNet-34-20 & 85.14 & \underline{43.98} & \bf{45.92} & 1.94\\
\cite{Huang2021} & WideResNet-34-R & 90.56 & \underline{36.91} & \bf{38.04} & 1.13\\
\cite{Huang2021} & WideResNet-34-R & 91.23 & \underline{35.91} & \bf{36.93} & 1.02\\
\cite{Gowal2020} & WideResNet-70-16 & 91.10 & \underline{33.33} & \bf{33.91} & 0.58\\
\cite{Gowal2020} & WideResNet-70-16 & 85.29 & \underline{42.04} & \bf{42.59} & 0.55\\
\cite{Gowal2021} & WideResNet-70-16 & 88.74 & \underline{32.08} & \bf{33.45} & 1.37\\
\cite{Rebuffi2021} & WideResNet-70-16 & 88.54 & \underline{35.02} & \bf{35.54} & 0.52\\
\cite{Rebuffi2021} & WideResNet-70-16 & 92.23 & \underline{32.40} & \bf{33.13} & 0.73\\
\toprule
ImageNet ($\varepsilon=4/255$) &&&&\\ \hline\hline
\cite{engstrom2019} & ResNet-50 & 62.56 &\underline{67.36} & \bf{69.58} & 2.22\\
\cite{DBLP:journals/corr/abs-2007-08489} & ResNet-18 & 52.92 &\underline{72.78} & \bf{74.34} & 1.56\\
\cite{DBLP:journals/corr/abs-2007-08489} & WideResNet-50-2 & 68.46 &\underline{58.38} & \bf{60.90} & 2.52\\
\cite{Wong2020} & ResNet-50 & 55.62 &\underline{71.38} & \bf{73.00} & 1.62\\
\cite{DBLP:journals/corr/abs-2007-08489} & ResNet-50 & 64.02 &\underline{62.40} & \bf{64.70} & 2.30\\
\hline\hline
\end{tabular}
}
\vskip -1.0in
\end{table*}

\begin{table*}
\centering
\caption{
Comparison of the ASR of APGD and ACG with DLR loss as the objective function.
Dataset: CIFAR-100($\varepsilon=\frac{8}{255}$)
}
\label{table:dlr_center_random_cifar100}
\vskip 0.15in
\begin{tabular}{ c | c | c | cc || c}
\toprule
CIFAR-100 ($\varepsilon=8/255$) & & & \multicolumn{2}{c}{Attack Success Rate} \\ \hline
\bf{paper} & \bf{Architecture} & \bf{clean acc} & \bf{APGD(5)} & \bf{ACG(5)} & \bf{diff} \\ \hline\hline
\cite{Rade2021} & PreActResNet-18 & 61.50 &\underline{70.52} & \bf{71.03} & 0.51\\
\cite{Wu2020a} & WideResNet-34-10 & 60.38 & \underline{68.97} & \bf{70.64} & 1.67\\
\cite{Rebuffi2021} & WideResNet-28-10 & 62.41 &\underline{66.64} & \bf{67.71} & 1.07\\
\cite{Rebuffi2021} & WideResNet-70-16 & 63.56 &\underline{64.14} & \bf{65.07} & 0.93\\
\cite{chen2020Efficient} & WideResNet-34-10 & 62.14 &\underline{71.77} & \bf{72.50} & 0.73\\
\cite{Chen2021} & WideResNet-34-10 & 64.07 &\underline{68.31} & \bf{69.11} & 0.80\\
\cite{Rice2020} & PreActResNet-18 & 53.83 &\underline{79.83} & \bf{80.76} & 0.93\\
\cite{Hendrycks2019} & WideResNet-28-10 & 59.23 &\underline{68.37} & \bf{70.73} & 2.36\\
\cite{Cui2020} & WideResNet-34-20 & 62.55 &\underline{67.78} & \bf{69.47} & 1.69\\
\cite{Rebuffi2021} & PreActResNet-18 & 56.87 &\underline{70.86} & \bf{71.42} & 0.56\\
\cite{Sitawarin2020} & WideResNet-34-10 & 62.82 &\underline{72.83} & \bf{74.93} & 2.10\\
\cite{Addepalli2021} & WideResNet-34-10 & 65.73 &\underline{68.65} & \bf{68.80} & 0.15\\
\cite{Cui2020} & WideResNet-34-10 & 60.64 &\underline{70.92} & \bf{72.89} & 1.97\\
\cite{Gowal2020} & WideResNet-70-16 & 60.86 &\underline{69.01} & \bf{69.85} & 0.84\\
\cite{Gowal2020} & WideResNet-70-16 & 69.15 &\underline{61.39} & \bf{62.73} & 1.34\\
\cite{Addepalli2021} & PreActResNet-18 & 62.02 &\underline{72.19} & \bf{72.58} & 0.39\\
\cite{Cui2020} & WideResNet-34-10 & 70.25 &\underline{69.35} & \bf{70.72} & 1.37\\
\hline\hline
\end{tabular}
\vskip -0.1in
\end{table*}

\begin{table*}[tb]
\centering
\caption{
The experimental result of the representative seven formulas to calculate $\beta$ for ACG.
}
\vskip 0.15in
\begin{tabular}{ c | c | ccccccc}\toprule
CIFAR-10 ($\varepsilon=8/255$) & & \multicolumn{7}{c}{Attack Success Rate}\\ \midrule
\bf{paper} & \bf{Architecture} & \bf{FR} & \bf{PR} & \bf{HS}  &  \bf{DY} &  \bf{HZ} &  \bf{DL} &  \bf{LS} \\ 
 \hline\hline
 \cite{Ding2019}&WideResNet-28-4 & 48.88& \underline{52.78}&\textbf{55.77} & 48.08& 49.98&44.87&52.05 \\
 \cite{Carmon2019}& WideResNet-28-10& 39.03& 39.55&\textbf{40.03} & 35.43& 39.05& 33.70 &\underline{39.56}\\
 \cite{Rebuffi2021}&PreActResNet-18 & 42.75& 42.88&\textbf{43.15} & 40.68& 42.50&40.20&\underline{42.90} \\
\bottomrule
\end{tabular}
\vskip -0.1in
\label{appendix:table:other_formula_result}
\end{table*}
\section{Experiment Environments}
\label{appendix:experiment_environments}
\begin{table*}
    \centering
    \caption{
    Computational environments: Information on the five computers used in the experiments are shown.
    }

    \label{tab:machine_environment}
    \vskip 0.15in
    \begin{tabular}{c|c|c|c}
    \toprule
        Machine & No.1 \& No.2 & No.3 \& No.4 &  No.5 \\
        \midrule
        \textrm{CPU} & \begin{tabular}{c} Intel(R) Xeon(R) Gold 6240R \\ CPU @ 2.40GHz $\times 2 $ \end{tabular} & \begin{tabular}{c} Intel(R) Xeon(R) Silver 4216 \\ CPU @ 2.10GHz $\times 2$ \end{tabular} & \begin{tabular}{c} Intel(R) Xeon(R) Gold 5120 \\ CPU @ 2.20GHz $\times 2$ \end{tabular}\\
        \hline
        \textrm{GPU} & \multicolumn{3}{c}{\textrm{NVIDIA GeForce RTX 3090} $ \times 4$}  \\
        \hline
        \textrm{RAM} & \multicolumn{2}{c|}{768\textrm{GB}} & 256\textrm{GB}\\
        \bottomrule
    \end{tabular}
    \vskip -0.1in
\end{table*}
The computational environments for our experiments, such as the CPU and GPU specifications and RAM capacity, are provided in \Cref{tab:machine_environment}. More information is also provided in the source codes.

\section{Experimental results of the evaluation of the representative $\beta$ formulas}
\label{appendix:OtherConjugateFormulation}
In this section, among the formulas used to determine $\beta$ proposed in prior works, we verified the effectiveness of seven representative formulas are verified through secondary experiments. Based on these experiments, we chose the formula for our approach. The formulas we verified in this experiment are given as follows.

\begin{itemize}
    \item FR: $\displaystyle \iter{\beta_{FR}}{k} = \frac{\| \nabla f(\iter{\bs{x}}{k}) \|_2^2}{\|  f(\iter{\bs{x}}{k-1})\|_2^2}$.
    \item PR: $\displaystyle \iter{\beta_{PR}}{k} = \frac{\langle \nabla f(\iter{\bs{x}}{k}), \iter{\bs{y}}{k-1} \rangle}
{\|  f(\iter{\bs{x}}{k-1}) \|_2^2}$.
    \item HS: $\displaystyle \iter{\beta_{HS}}{k} = -\frac{\langle \nabla f(\iter{\bs x}{k}), \iter{\bs{y}}{k-1}\rangle}
{\langle \iter{\bs s}{k-1}, \iter{\bs y}{k-1}\rangle }$.
    \item DY: $\displaystyle \iter{\beta_{DY}}{k} = \frac{\| \nabla f(\iter{\bs{x}}{k}) \|_2^2}
{\langle \iter{\bs s}{k-1}, \iter{\bs y}{k-1}\rangle }$.
    \item HZ: $\displaystyle \iter{\beta_{HZ}}{k} = \frac{
    \left\langle
    \iter{\bs{y}}{k-1} - 
    \displaystyle
    \frac{
        2 \iter{\bs{s}}{k-1} \| \iter{\bs{y}}{k-1} \|_2^2
        }{
        \langle \iter{\bs{s}}{k-1}, \iter{\bs{y}}{k-1} \rangle
    }
    ,
    \nabla f ( \iter{\bs{x}}{k}) 
    \right\rangle
    }{
      \langle \iter{\bs s}{k-1}, \iter{\bs y}{k-1}\rangle 
    }
    $.
    \item DL: 
        $\displaystyle \iter{\beta_{DL}}{k} = 
        - \frac{
            \left \langle \iter{\bs{y}}{k-1} - t \iter{\bs{s}}{k-1}, \nabla f (\iter{\bs{x}}{k}) \right\rangle 
        }{
        \langle \iter{\bs{s}}{k-1}, \iter{\bs{y}}{k-1} \rangle
        }, \ t \geq 0
        $.
    \item LS: 
    $\displaystyle \iter{\beta_{LS}}{k} = 
    - \frac{
        \langle \nabla f(\iter{\bs{x}}{k}), \iter{\bs{y}}{k-1} \rangle
    }{
        \left\langle \iter{\bs{s}}{k-1}, \nabla f (\iter{\bs{x}}{k-1}) \right \rangle
    }
    $.
\end{itemize}

We conducted small experiments on only models described in \cite{Ding2019,Carmon2019,Rebuffi2021} using the same experimental setup described in \Cref{sec:experiments}. The CIFAR-10 dataset was used, and the results are presented in \Cref{appendix:table:other_formula_result}. \Cref{appendix:table:other_formula_result} shows that the ASR of the formulas whose enumerator is $\langle\nabla f(\bs{x}), \bs{y} \rangle$ (PR, HS, LS) were higher than those of the formulas whose numerator is $\|\nabla f(\bs{x})\|_2^2$ (FR, DY), and that the ASR of the HS formula was the highest.

\section{Analysis of WideResNet-34-10 ~\cite{Addepalli2021}}
\label{appendix:addepalli_analysis}
\begin{figure}[tb]
    \vskip 0.2in
    \centering
    \includegraphics[width=\linewidth]{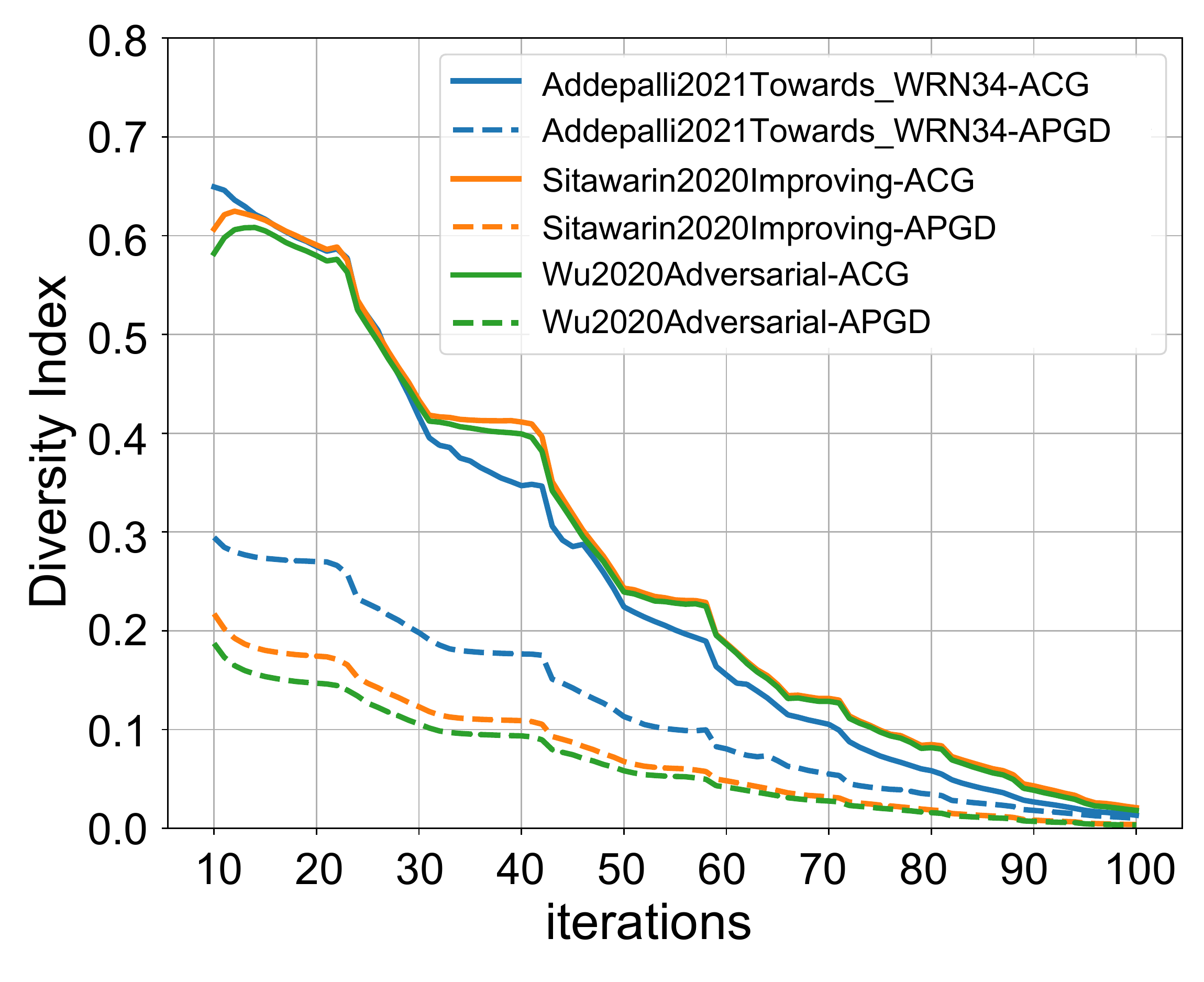}
    \caption{
    \measurement of ACG and APGD on three models \cite{Addepalli2021,Wu2020a,Sitawarin2020}
    }
    \label{figure:addepalli_cifar100}
    \vskip -0.2in
\end{figure}

We analyzed the search behavior of ACG on a WideResNet-34-10 model trained by the method proposed in \cite{Addepalli2021} using CIFAR-100, the only model in which the ASR of ACG was lower than that of APGD in the experiment using the CW loss (see \Cref{table:cifar100_large}).
From \Cref{figure:addepalli_cifar100}, it may be observed that the \measurement of ACG of this model was lower than that of ACG on the other models. 
This means that the search for ACG in this model tended to be more intensified than the attacks on the other two models \cite{Sitawarin2020,Wu2020a}.
By contrast, the \measurement of APGD on this model was higher than that of APGD in the other models, indicating that it tended to be more diversified.
These results suggest that the reason why APGD was superior to ACG for this model was because intensification was required to achieve a higher ASR.
However, the models requiring intensification for more effective attacks were rare because ACG exhibited a higher ASR than APGD for most of the models in \Cref{table:cifar100_large,table:cifar10_large,table:imagenet_large}.

\section{Generalization of \measurement }
\label{generalized_DI}
In this section, we generalize the \measurement defined in \Cref{subsection:measurement} to arbitrary bounded distance spaces.
First, we describe the definition of local and global clustering coefficients for undirected graphs and then use the global clustering coefficient to define the \measurement for arbitrarily bounded distance spaces. Then, using the global clustering coefficient, we define the \measurement for arbitrary bounded distance spaces.

\subsection{Definition of the local clustering coefficient \cite{Kemper2010}}
\label{appendix:clustering_coefficient}
Let $G=(V, E)$ be an undirected graph. Let the set of neighbor nodes of node $\mathfrak{v}_i\in V$ be $\mathcal{N}_i:=\{\mathfrak{v}_j\in V\mid e_{ij}\in E\}$. Then the local clustering coefficient $C_{\mathfrak{v}_i}(G)$ at node $\mathfrak{v}_i$ of graph $G$ is defined as follows.
\begin{equation}
\label{local_clustering_coef}
C_{\mathfrak{v}_i}(G):=\frac{2\#\{e_{jk}\in E\mid e_{ij}\in E\land e_{ik}\in E\}}{\#\mathcal{N}_i\times(\#\mathcal{N}_i-1)}
\end{equation}
According to this definition, $C_{\mathfrak{v}_i}(G)\in [0,1]$.

\subsection{Definition of the global clustering coefficient \cite{Kemper2010}}
Let $G=(V, E)$ be an undirected graph, and the global clustering coefficient $C(G)$ of the graph $G$ be defined using the equation \eqref{local_clustering_coef} as follows.
\begin{equation}
    C(G):= \frac{1}{\# V}\sum_{\mathfrak{v}\in V}C_\mathfrak{v}(G)
\end{equation}
In the same manner as $C_{\mathfrak{v}_i}(G)$, $C(G)$ also satisfies $0\leq C(G)\leq 1$.

\subsection{Definition of \measurement on the general bounded metric spaces}
Let $(U, d)$ be a bounded distance space, $\mathcal{V}\subset U$ be its finite subset, and $\theta\in\R$ be the graph of $G_{\mathcal{V}}(\theta):=(\mathcal{V}, E(\theta))$, where $E(\theta)=\{(v, w)\in \mathcal{V}\times \mathcal{V}\mid d(v, w) \leq \theta\}$. In this case, as $M=\sup_{v,w\in U}d(v,w)$, \measurement is defined as follows.
\begin{equation}
    \textrm{\measurement}(\mathcal{V}, M):=1-\frac{1}{M}\int_{0}^{M}C(G_\mathcal{V}(\theta))d\theta
\end{equation}

\section{Examples of \measurement}
\label{additional_examples_of_DI}
\begin{figure*}[tb]
  \centering
  \begin{tabular}{c} 
      \includegraphics[valign=m,width=0.8\linewidth]{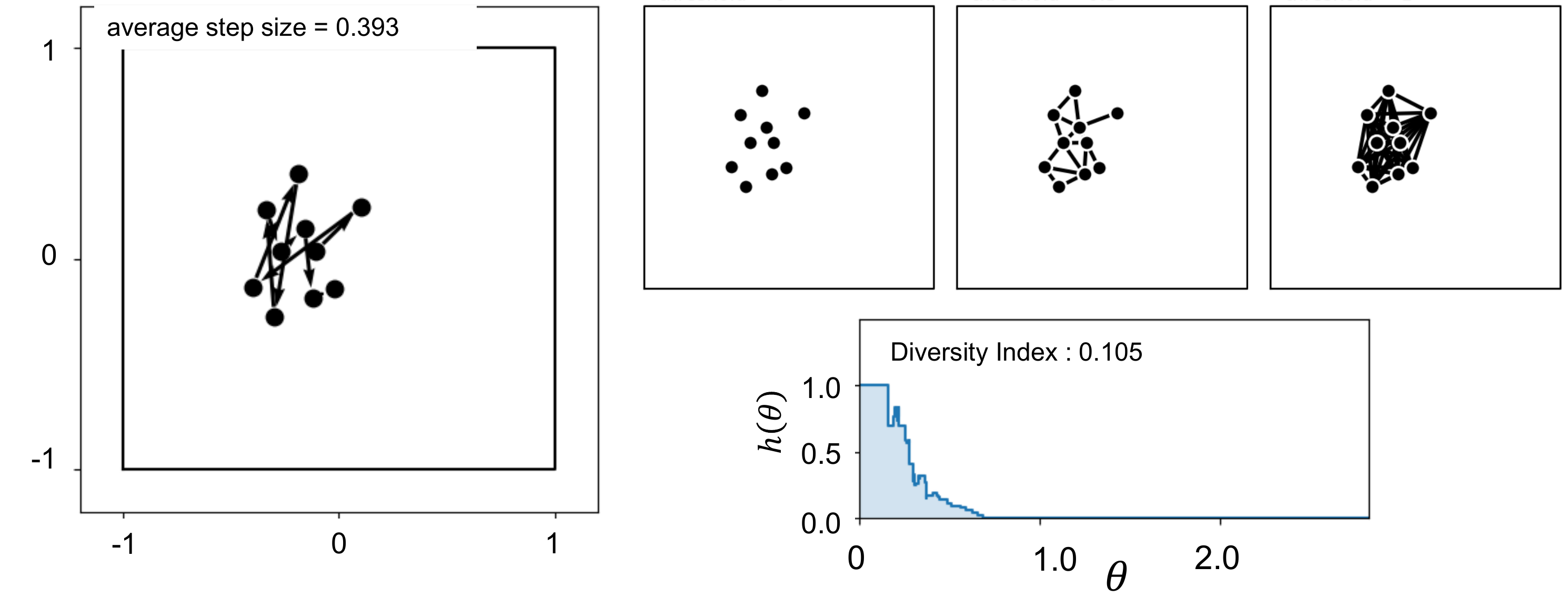}
    \end{tabular}
    \begin{tabular}{c}
    \hline
      \includegraphics[valign=m,width=0.8\linewidth]{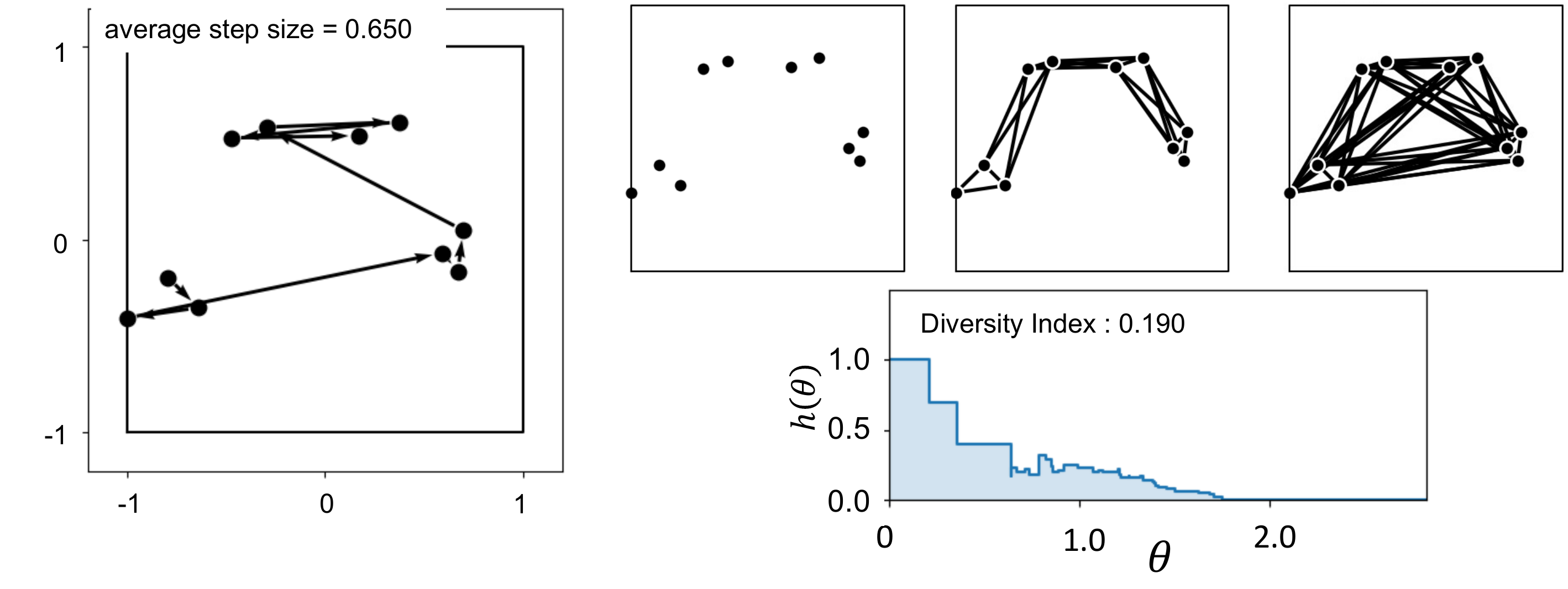}
    \end{tabular}
    \begin{tabular}{c} 
    \hline 
      \includegraphics[valign=m,width=0.8\linewidth]{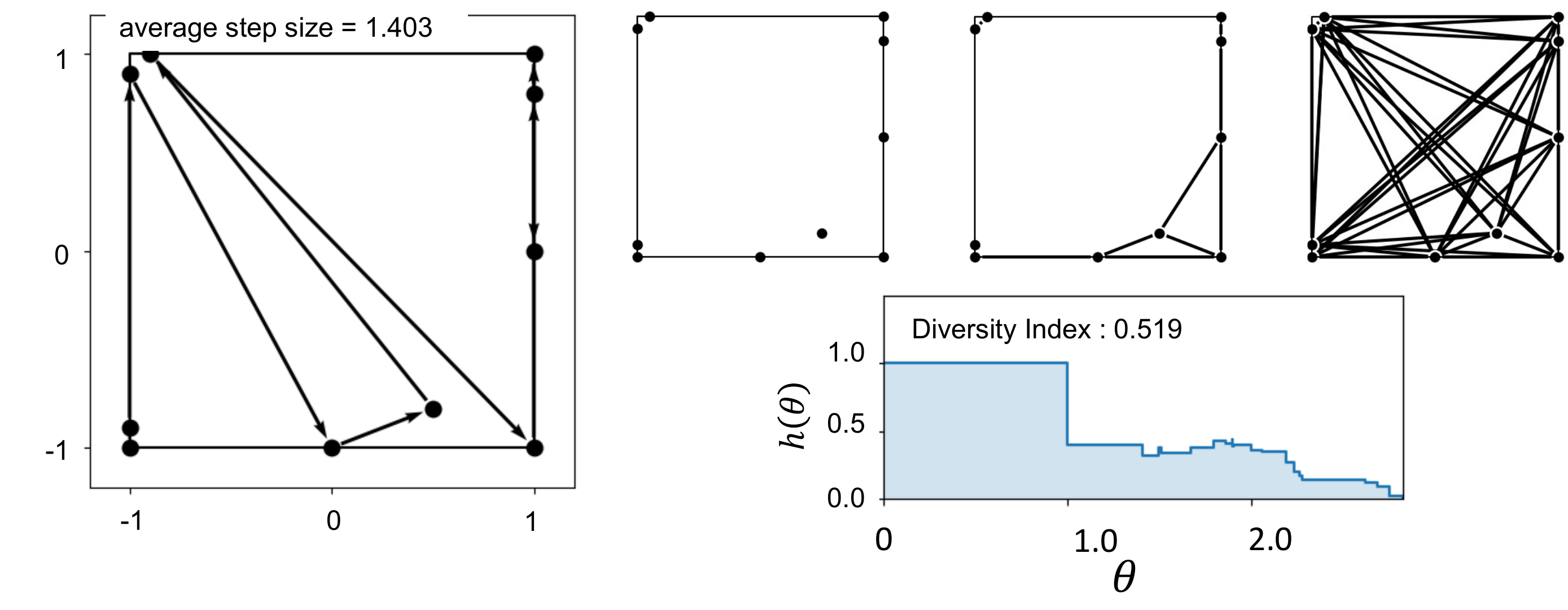}
     \end{tabular}
   \caption{
   Additional examples of point clouds and its {\measurementNoSpace}s.
   }
   \label{extra_examples_of_DI}
\end{figure*}

\Cref{extra_examples_of_DI} shows an example of the calculation of \measurement for point clouds with different distributions.
The first row of \Cref{extra_examples_of_DI} shows an example of a point cloud and its \measurement in which the point clouds form a single cluster. The second row shows an example of a point cloud and its \measurement, in which the point clouds form approximately three to four clusters. The third row shows an example of a point cloud and its \measurement where most points are distributed on the boundary. The final example is diverse in that it shows  a diversity search performed on the boundary.
These examples show that \measurement takes a small value when the point cloud is dense or when the clusters are formed, and \measurement takes a relatively large value when there are no clusters the elements of which number greater than 2.

\section{Evaluating the performance of ACG combined with Auto Attack}
We did not compare the ASR of Auto Attack(AA) and ACG directly because we focused on generating many adversarial examples quickly. AA comprises of four different algorithms for adversarial attacks, and it takes much longer than ACG. 
We considered that evaluating ACG as a component of AA would be more suitable; we therefore constructed AA(ACG-CE) using ACG with untargeted cross-entropy loss instead of APGD with the same loss. 
From the numerical results shown in \Cref{table:compare_asr} below, it may be observed that ACG is also a useful as a component of AA.
We expect that future work along there lines will adopt an appropriate combination of existing algorithms, including ACG.

\begin{table*}[tb]
\centering
\caption{
    The ASR of AutoAttack(Reported) and AutoAttack combined with ACG with CE loss (AA(ACG-CE)).
}
\label{table:compare_asr}
\begin{tabular}{c|c|cc}
\toprule
\bf{paper} & \bf{Architecture} & AA(ACG-CE) & Reported \\
\midrule
\cite{Ding2019} & WideResNet-28-4 & \bf{58.60} & 58.56\\
\cite{Carmon2019}& WideResNet-28-10 & \bf{40.48} & 40.47\\
\cite{andriushchenko2020square}& WideResNet-18 & \bf{56.07} & \bf{56.07}\\
\cite{Rebuffi2021}& PreActResNet-18 & \bf{43.34} & \bf{43.34}\\
\midrule
CIFAR 100 &&&\\
\cite{Rice2020}& PreActResNet-18 & 81.02 & \bf{81.05}\\
\midrule
ImageNet &&&\\
\cite{engstrom2019}& ResNet-50 & 70.76 & \bf{70.78}\\
\cite{Wong2020}& ResNet-50 &  \bf{73.80} & 73.76\\
\cite{DBLP:journals/corr/abs-2007-08489}& ResNet-50 & \bf{65.36} & 65.04\\
\bottomrule
\end{tabular}
\end{table*}

\end{subappendices}

\end{document}